\documentclass{www2013-annotated}

\usepackage[ruled,vlined,linesnumbered]{algorithm2e}
\usepackage{algpseudocode}
\usepackage{amsmath}
\usepackage{amssymb}
\usepackage{bm}
\usepackage{bbding}
\usepackage{caption}
\usepackage{color}
\usepackage{enumitem}
\usepackage{epsfig}
\usepackage{subfigure}
\usepackage{tikz}
\usepackage{verbatim}

\usetikzlibrary{arrows,calc,shapes,snakes,automata,backgrounds,fit,petri}

\def\figures{0}

\def\pointer{\HandRight $\phantom{0}$}

\def\usr{\mathop{\rm user}}
\def\itm{\mathop{\rm item}}

\def\Ebb{\mathbb{E}}
\def\Ibb{\mathbb{I}}
\def\Rbb{\mathbb{R}}

\def\bmu{\bm{\mu}}

\def\bpi{\bm{\pi}}

\def\btau{\bm{\tau}}
\def\btheta{\bm{\theta}}

\def\bxi{\bm{\xi}}

\def\0{\mathbf{0}}

\def\b{\mathbf{b}}

\def\u{\mathbf{u}}
\def\v{\mathbf{v}}
\def\x{\mathbf{x}}

\def\z{\mathbf{z}}

\def\G{\mathbf{G}}
\def\H{\mathbf{H}}
\def\I{\mathbf{I}}

\def\P{\mathbf{P}}

\def\S{\mathbf{S}}
\def\U{\mathbf{U}}
\def\V{\mathbf{V}}

\def\X{\mathbf{X}}

\def\<{ \left\langle }
\def\>{ \right\rangle }

\def\Bcal{\mathcal{B}}

\def\Gcal{\mathcal{G}}
\def\Hcal{\mathcal{H}}
\def\Lcal{\mathcal{L}}
\def\Mcal{\mathcal{M}}
\def\Ncal{\mathcal{N}}
\def\Ocal{\mathcal{O}}

\def\wo{\backslash}
\def\defined{\stackrel{\text{\tiny def}}{=}}

\definecolor{Blue}{rgb}{0.0,0.0,1.0}
\newcommand{\bluetext}[1]{{\color{Blue}{#1}}}

\definecolor{White}{rgb}{1.0,1.0,1.0}
\newcommand{\whitetext}[1]{{\color{White}{#1}}}

\begin{document}

\title{One-class Collaborative Filtering with Random Graphs: \bluetext{Annotated Version}}

\numberofauthors{2}
\author{
%
%
\alignauthor
Ulrich Paquet \\
       \affaddr{Microsoft Research}\\
       \email{ulripa@microsoft.com}
\alignauthor
Noam Koenigstein \\
       \affaddr{Microsoft}\\
       \email{noamko@microsoft.com}
}

\maketitle
\begin{abstract}
The bane of one-class collaborative filtering is interpreting and modelling the latent signal from the missing class.
In this paper we present a novel Bayesian generative model for implicit collaborative filtering.
It forms a core component of the Xbox Live architecture, and unlike previous approaches, delineates the odds of a user disliking an item from simply not considering it.
The latent signal is treated as an unobserved random graph connecting users with items they might have encountered.
We demonstrate how large-scale distributed learning can be achieved through a combination of stochastic gradient descent and mean field variational inference over random graph samples.
A fine-grained comparison is done against a state of the art baseline on real world data.
\end{abstract}




\section{Introduction}

This paper highlights a solution to a very specific problem, the prediction of a ``like'' or ``association'' signal from one-class data. One-class or ``implicit'' data surfaces in many of Xbox's verticals, for example when users watch movies through Xbox Live.
In this vertical, we recommend media items to users, drawing on the correlations of their viewing patterns with those of other users.
We assume that users don't watch movies that they dislike; therefore the negative class is absent.
The problem is equivalent to predicting new connections in a network: given a disjoint user and an item vertex, what is the chance that they should be linked?

We introduce a Bayesian generative process for connecting users and items. It models the ``like'' probability by interpreting the missing signal as a two-stage process: firstly, by modelling the odds of a user considering an item, and secondly, by eliciting a probability that that item will be viewed or liked.
This forms a core component of the Xbox Live architecture, serving recommendations to more than 50 million users worldwide, and replaces an earlier version of our system \cite{XboxPaper}.
The two-stage delineation of popularity and personalization allows systems like ours to trade them off in optimizing user-facing utility functions. 
The model is simple to interpret, allows us to estimate parameter uncertainty, and most importantly, easily lends itself to large-scale inference procedures.

Interaction patterns on live systems typically follow a power-law distribution, where some users or items are exponentially more active or popular than others. We base our inference on a simple assumption, that the missing signal should have the same power-law degree distribution as the observed user-item graph. Under this assumption, we learn latent parametric descriptions for users and items by computing statistical averages over all plausible ``negative graphs''.

The challenge for one-class collaborative filtering is to treat the absent signal without incurring a prohibitive algorithmic cost. Unlike its fully binary cousin, which observes ``dislike'' signals for a selection of user-item pairs, each unobserved user-item pair or edge has a possible negative explanation. For $M$ users and $N$ items, this means that inference algorithms have to consider $\Ocal(MN)$ possible negative observations. In problems considered by Xbox, this amounts to modelling around $10^{12}$ latent explanations. The magnitude of real world problems therefore casts a shadow on models that treat each absent observation individually \cite{Palla_2012}.

Thus far, solutions to large-scale one-class problems have been based on one of two main lines of thought.
One line formulates the problem as an objective function over all observed and missing data, in which the contribution by the ``missing data'' drops out in the optimization scheme \cite{Hu_2008}. 
It relies on the careful assignment of confidence weights to all edges, but there is no methodical procedure for choosing these confidence weights except an expensive exhaustive search via cross-validation.
If a parametric definition of confidence weights is given, a low rank approximation of the weighting scheme can also be included in an objective function \cite{Pan_2009}.
The work presented here differs from these approaches
by formulating a probabilistic model rather than an optimization problem,
and quantifies our uncertainty about the parameters and predictions.

A second approach is to randomly synthesize negative examples.
Our work falls in this camp, for which there already exists a small body of work. The foremost of these is arguably Bayesian Personalized Ranking (BPR), which converts the one-class problem into a ranking problem \cite{Rendle_2009}. In it, it is assumed that the user likes everything that she has seen more than the items that she hasn't seen. 
This assumption implies a constrained ordering of many unobserved variables, one arising from each item.
This user-wise ranking of items facilitates the inference of latent features for each user and item vertex.
By design, there is no distinction between missing items in BPR; however, popularity sampling of the unobserved items was employed to give more significance to popular missing items \cite{Zeno_BPR_JMLR}. This approach was effectively utilized by many of the leading solutions in the KDD-Cup'11 competition \cite{KDDCup11}.
An alternative, more expensive approach is to construct an ensemble of solutions, each of which is learned using a different sample of synthesized ``negative'' edges \cite{Pan_2008}.

We motivate our approach by discussing properties of typical bipartite real world graphs in Section \ref{sec:realworld}.
A generative model for collaborative filtering when such graphs are observed is given in Section \ref{sec:coll}.
A component of the model is the hidden graph of edges---items that a user considered, but didn't like.
Section \ref{sec:randomgraphs} addresses the hidden graph as a random graph.
Section \ref{sec:variational} combines variational inference and stochastic gradient descent to present methods for large scale parallel inference for this probabilistic model.
In Section \ref{sec:results}, we show state of the art results on two practical problems, a sample of movies viewed by a few million users on Xbox consoles, and a binarized version of the Netflix competition data set.

\section{Typical Real World Data} \label{sec:realworld}

\begin{figure}[t]
\begin{center}
\ifnum\figures=1
\includegraphics[width=0.45\textwidth]{Results/Figures/XboxDegrees.eps}
\\ \vspace{5pt}
\includegraphics[width=0.45\textwidth]{Results/Figures/Netflix4and5Degrees.eps}
\else
\includegraphics[width=0.45\textwidth]{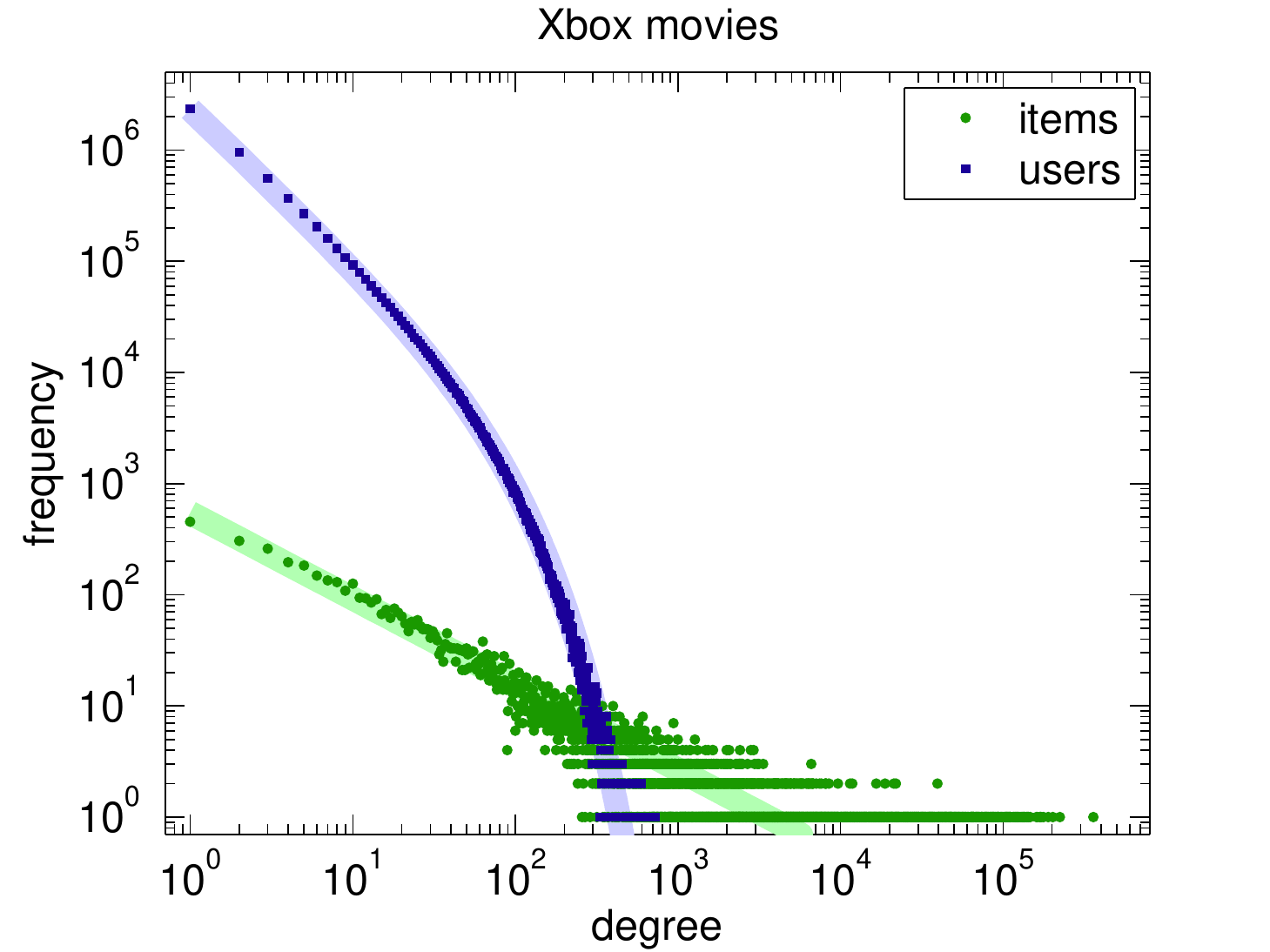}
\\ \vspace{5pt}
\includegraphics[width=0.45\textwidth]{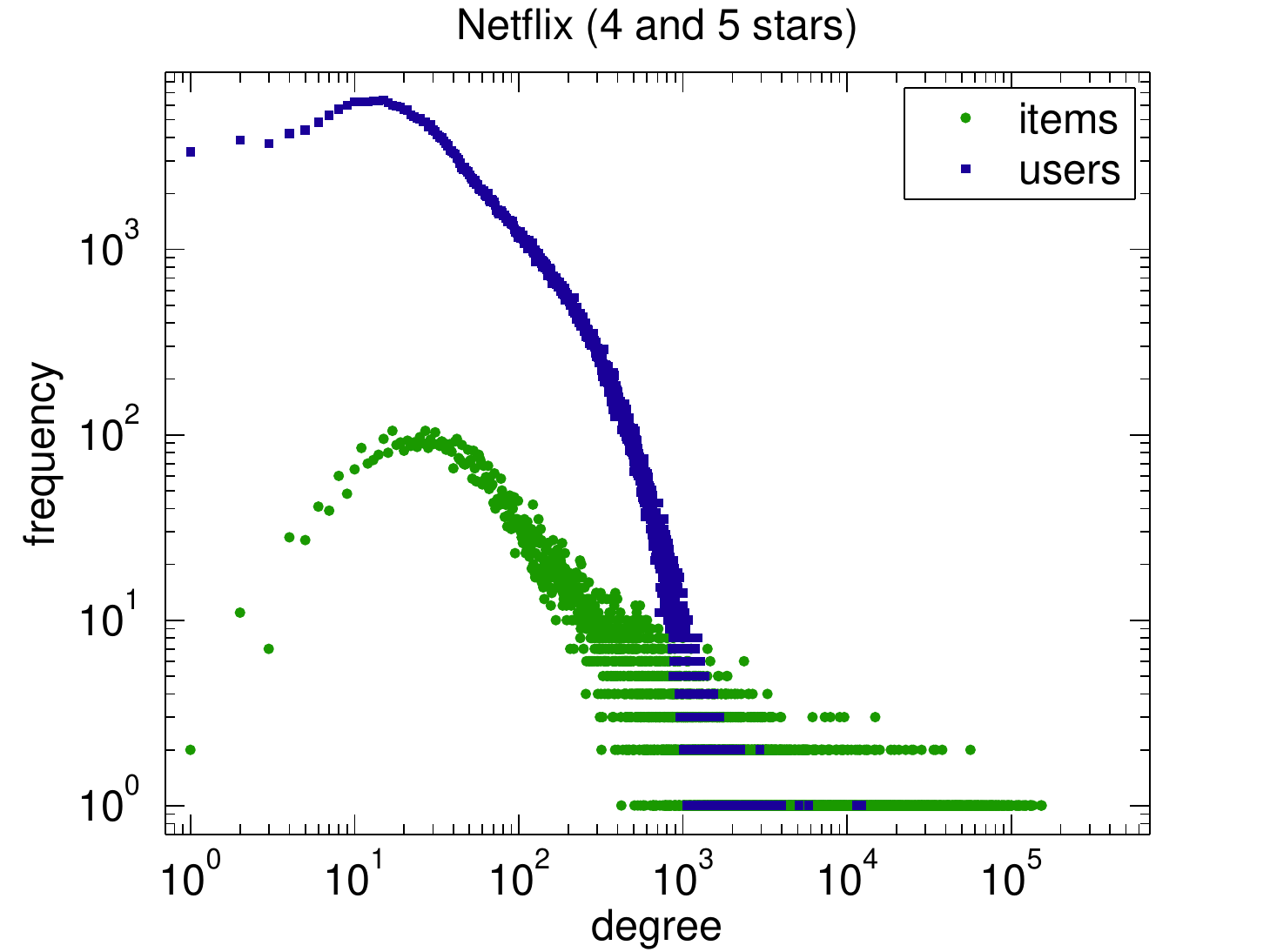}
\fi
\end{center}
\caption{Degree distributions for two bipartite graphs between users and movies: a sample of $4.4 \times 10^{7}$ edges for movies viewed on Xbox \emph{(top)} and the $5.6 \times 10^{7}$ four and five starred edges in the Netflix prize data set \emph{(bottom)}. }
\label{fig:degreedistributions}
\end{figure}

The frequency of real-world interactions typically follows some form of power-law.
In Xbox Live, we observe a bipartite graph $\G$ of $M$ users and $N$ items, with two (possibly vastly) different degree distributions for the two kinds of vertices. Figure \ref{fig:degreedistributions} \emph{(top)} illustrates the degree distribution of a sample of $M = 6.2 \times 10^6$ users that watched $N = 1.2 \times 10^4$ different movies on their Xbox consoles, where an edge appears if a user viewed a movie.
Throughout the paper the edges in the \emph{observed} graph $\G$ will be denoted with the binary variable $g_{mn} \in \{ 0, 1 \}$ for vertices $m$ and $n$, with a zero value indicating the absence of an edge.
We denote the observed degree distributions as $p_{\usr}(d)$ and $p_{\itm}(d)$. If a user viewed on average $\mu$ items, and an item was viewed on average $\nu$ times, then $\Ebb_{p_{\usr}}[d] = \mu$ and $\Ebb_{p_{\itm}}[d] = \nu$, and the constraint
\begin{equation} \label{eq:consistency}
\frac{\mu}{N} = \frac{\nu}{M}
\end{equation}
should hold \cite{Newman_2001}.
In Figure \ref{fig:degreedistributions} \emph{(top)}, the empirical distributions satisfy $\mu = 7.1$ and $\nu = 3780$, validating the mean constraint
$\mu / N = \nu / M = 0.0006$. We overlay a power law degree distribution to items $p_{\itm}(d) \propto d^{-0.77}$. The user distribution exhibits an marked exponential cut-off, with $p_{\usr}(d) \propto d^{-1.4} \, \mathrm{e}^{-d / 70}$, and shares its form with many scientific collaboration networks \cite{Newman_2002}.
The degree distribution of the publicly available Netflix data set is shown in Figure \ref{fig:degreedistributions} \emph{(bottom)}. 
In it, we have $M = 4.8 \times 10^5$ users and $N = 1.8 \times 10^4$ items. 
We took a positive edge to be present if a user rated an item with four or five stars. 

Given $p_{\usr}(d)$ and $p_{\itm}(d)$, one can sample i.i.d.~graphs with the given degree distribution. Firstly, generate vertex degrees for each user and item at random, and calculate their sum.
If the sums are unequal, randomly choose one user and item, discard their degrees, and replace them with new degrees of the relevant distributions. This process is repeated until the total user and item degrees are equal, after which vertex pairs are randomly joined up \cite{Newman_2001}.

\section{Collaborative filtering} \label{sec:coll}

Our collaborative filtering model rests on a basic assumption, that if an edge $g_{mn} = 1$ appears in $\G$, user $m$ liked item $n$.
However, a user must have considered additional items that she didn't like, even though the dislike or ``negative'' signals are not observed. This hidden graph with edges $h_{mn} \in \{0,1\}$ is denoted by $\H$. We say that a user considered an item if and only if $h_{mn}=1$, and the rule $g_{mn} = 1 \Rightarrow h_{mn} = 1$ holds; namely, a user must have considered all the items that she ``liked'' in $\G$.
The latent signal is necessary in order to avoid trivial solutions, where the interpretation inferred from data tells us that everyone likes everything or that every edge should be present. It strongly depends on our prior beliefs about $\H$, like its degree distribution or power-law characteristics.
$\G$ is observed as a subgraph of $\H$, while
the rest of the edges of the hidden graph $\H$ form the unobserved ``negative'' signals.

\subsection{The likelihood and its properties}

On knowing the hidden graph, we define a bilinear or ``matrix factorization'' collaborative filtering model.
We associate a latent feature $\u_m \in \Rbb^K$ with each user vertex $m$, and $\v_n \in \Rbb^K$ with each item vertex $n$. Additionally, we add biases $b_m \in \Rbb$ and $b_n \in \Rbb$ to each user and item vertex.
The odds of a user liking or disliking an item under consideration ($h = 1$) is modelled with
\begin{equation} \label{eq:g}
p(g \, | \, \u, \v, b, h = 1) = \sigma\big( \u^T \v + b \big)^{g}
\Big[1 - \sigma\big( \u^T \v + b \big)\Big]^{1 - g} \ ,
\end{equation}
with the logistic or sigmoid function being $\sigma(a) = 1 / (1 + \mathrm{e}^{-a})$, with $a \defined \u^T \v + b$.
Subscripts $m$ and $n$ are dropped in (\ref{eq:g}) as they are clear from the context; $b$ denotes the sum of the biases $b_m + b_n$.
The likelihood of $g$ for any $h$ is given by the expression
\begin{equation} \label{eq:singlelikelihood}
p(g \, | \, a, h) = \big[ \sigma(a)^{g}
(1 - \sigma(a))^{1 - g} \big]^{h} \cdot (1 - g)^{1 - h} \ .
\end{equation}
As $g = 1 \Rightarrow h = 1$ by construction, the last factor can be ignored in (\ref{eq:singlelikelihood}). If the binary ``considered'' variable $h$ is marginalized out in (\ref{eq:singlelikelihood}), we find that
\begin{align}
p(g  = 1 \, | \, a) & = p(h = 1) \, \sigma(a) \ , \nonumber \\
p(g  = 0 \, | \, a) & =p(h = 1) (1 - \sigma(a)) + (1 - p(h = 1)) \ . \label{eq:seeingr}
\end{align}
In other words, the odds of encountering an edge in $\G$ is the product of two probabilities, separating \emph{popularity} from \emph{personalization}: $p(h = 1)$, the user considering an item, and $\sigma(a)$, the user then liking that item.

\subsection{The full model}

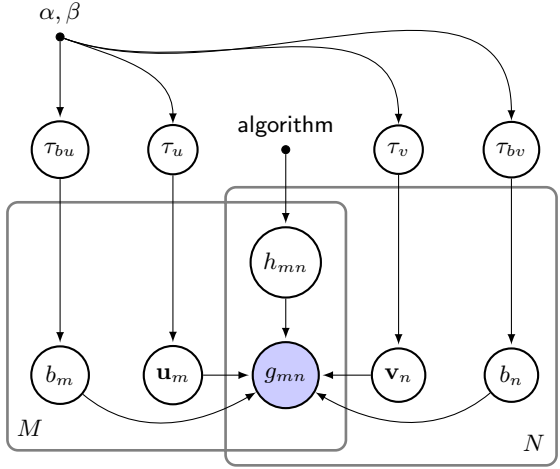
\begin{figure}[t]
\begin{center}
\begin{tikzpicture}[bend angle=45,>=latex]
\tikzstyle{obs} = [ circle, thick, draw = black!100, fill = blue!20, minimum size = 6mm ]
\tikzstyle{lat} = [ circle, thick, draw=black!100, fill = red!0, minimum size = 6mm ]
\tikzstyle{par} = [ circle, draw, fill = black!100, minimum width = 3pt, inner sep = 0pt]	
\tikzstyle{every label} = [black!100]

\begin{scope}[node distance = 1.5cm and 1.5cm,rounded corners=4pt]
\node [obs] (g) {$g_{mn}$}; 
\node [lat] (h) [ above of = g]  {$h_{mn}$}
      edge [post] (g);
\node [par] (r) [ above of = h, label = 90: ${\mathsf{algorithm}}$] {}
       edge [post] (h);
\node [lat] (u) [ left of = g ] {$\u_m$}
      edge [post] (g);
\node [lat] (bu) [ left of = u] {$b_m$};
\draw[-latex] (bu) to[out=320,in=210] (g);
\node [lat] (v) [ right of = g] {$\v_n$}
      edge [post] (g);
\node [lat] (bv) [ right of = v] {$b_n$};
\draw[-latex] (bv) to[out=220,in=330] (g);
\node [lat] (tauu) [ left of = r] {$\tau_u$}
      edge [post] (u);
\node [lat] (taubu) [ left of = tauu] {$\tau_{bu}$}
      edge [post] (bu);
\node [lat] (tauv) [ right of = r] {$\tau_v$}
      edge [post] (v);
\node [lat] (taubv) [ right of = tauv] {$\tau_{bv}$}
      edge [post] (bv);
\node [par] (ab) [ above of = taubu, label = 90: ${\alpha, \beta}$] {}
      edge [post] (taubu);
\draw[-latex] (ab) to[out=340,in=90] (tauu);
\draw[-latex] (ab) to[out=340,in=90] (tauv);
\draw[-latex] (ab) to[out=340,in=90] (taubv);
\draw (-3.4,-0.7) node {$M$};
\draw (3.3,-0.9) node {$N$};
				
\begin{pgfonlayer}{background}
\filldraw [line width = 1pt, draw=black!50, fill=white!100]
(-3.7cm,2.3cm) rectangle (0.8cm,-1.0cm)
(-0.8cm,2.5cm) rectangle (3.6cm,-1.2cm);
\end{pgfonlayer}							
\end{scope}
\end{tikzpicture}	
\end{center}
\caption{The graphical model for observing graph $\G$ connecting $M$ user with $N$ item vertices.
The prior on the hidden graph $\H$ is algorithmically determined to resemble the type of the observed graph.}
\label{fig:graphicalmodel}
\end{figure}

The probability of $\G$ depends on the prior distributions of the vertices' hidden features.
We choose them to be Gaussian: $p(\U) = \prod_{m=1}^{M} \Ncal(\u_m \, ; \, \0, \tau_u^{-1} \I)$  for the users, where $\U \defined \{ \u_m \}_{m=1}^{M}$,
with similar Gaussian priors on the parameters governing the item vertices.
These are shown in the graphical model in Figure \ref{fig:graphicalmodel}.
To infer the various scale parameters $\tau$, we place a conjugate Gamma hyperprior on each, for example
\[
{\Gcal}(\tau_u ; \alpha, \beta) = \beta^{\alpha} /  \Gamma(\alpha) \cdot \tau_u^{\alpha - 1}\mathrm{e}^{- \beta \tau_u} \ .
\]
The only prior beliefs in Figure \ref{fig:graphicalmodel} that do not take an explicit form is that of $\H$. It could be parameterized with a particular degree distribution, be it Poisson, exponential, or a power law with an exponential cut-off. 
However, we would like this to (approximately) be in the same family as the observed data, and determine an algorithm which can generate such graphs.
Section \ref{sec:randomgraphs} elaborates on this, including the closure of the graph family under random sampling of subnetworks.

We collectively denote our parameters by $\btheta \defined \{ \H, \U, \V, \b, \btau \}$,
with $b_{mn} \defined b_m + b_n$ as shorthand notation.
The joint density of all the random variables, given the hyperprior parameters $\alpha$ and $\beta$, is
\begin{align}
& p(\G, \btheta) = \prod_{m = 1}^{M} \prod_{n=1}^{N} \overbrace{
\sigma(\u_m^T \v_n + b_{mn})^{g_{mn}} \cdots \qquad \qquad \qquad}^{\ell_{mn}}
\nonumber \\
& \qquad\qquad\qquad\qquad \cdots 
\big[1 - \sigma(\u_m^T \v_n + b_{mn}) \big]^{h_{mn} (1 - g_{mn})} 
\nonumber \\
& \quad \cdot \prod_{m=1}^{M} \Ncal(\u_m ; \0, \tau_u^{-1} \I) \, \Ncal(b_m ; 0, \tau_{b_m}^{-1})
\cdot \Gcal(\tau_u ; \alpha, \beta)  \nonumber
\\
& \quad \cdot \prod_{n=1}^{N} \Ncal(\v_m ; \0, \tau_v^{-1} \I) \,
\Ncal(b_n ; 0, \tau_{b_n}^{-1}) \cdot \Gcal(\tau_v ; \alpha, \beta) \nonumber \\
& \quad \cdot \Gcal(\tau_{b_m} ; \alpha, \beta) \cdot \Gcal(\tau_{b_n} ; \alpha, \beta) \cdot p(\H) \ .
\label{eq:jointdensity}
\end{align}
The sigmoid product is denoted with $\ell_{mn}$, and will later appear in a variational bound in (\ref{eq:jjbound}).
Obtaining a posterior approximation to (\ref{eq:jointdensity}) would follow known literature \cite{Paquet_2012, Stern_2009}, were it not for the unknown occurrence of edges $h_{mn}$ in $\H$.
Sections \ref{sec:randomgraphs} and \ref{sec:variational} are devoted to treating $\H$.

One might also consider placing Normal-Wishart hyperprior on the means and variances of $\u_m$ and $\v_n$ \cite{Paquet_2012, Salakhutdinov_2008}. 
In practice, we benefit from additionally using meta-data features in the hyperpriors. They allow us to learn how shared features connect the prior distributions of various items, but is beyond the scope of this paper.

\subsection{Factorized approximation}

It is analytically intractable to compute the Bayesian averages necessary for marginalization in (\ref{eq:jointdensity}).
This hurdle is commonly addressed in one of two ways: Samples from the posterior can be drawn by simulating a Markov chain with the posterior as its stationary distribution, and these samples used for prediction \cite{Neal_1993}.
Alternatively, one might substitute the integration problems required for Bayesian marginalization with an optimization problem, that of finding the best deterministic approximation to the posterior density \cite{Jordan_1999}.

We \emph{approximate} the posterior from (\ref{eq:jointdensity}), rather than sample from it, as it allows a compact representation to be serialized to disk.
The posterior from (\ref{eq:jointdensity}) is approximated with the fully factorized distribution $q$,
\begin{align} 
p(\btheta | \G) \approx q(\btheta) & \defined \prod_{m=1}^{M} q(b_m) \prod_{k=1}^{K} q( u_{mk}) \cdot \prod_{n=1}^{N} q(b_n) \prod_{k=1}^{K} q(v_{nk}) \nonumber \\
& \quad \cdot q(\tau_u) \, q(\tau_v) \, q(\tau_{b_u}) \, q(\tau_{b_v}) \, q(\H) \ .
\label{eq:factorizing}
\end{align}
The factors approximating each of the vertex features in $\U$, $\V$, and $\b$ are chosen to be a Gaussian, for example $q(u_{mk}) = \Ncal( u_{mk} ;  \eta_{mk}, \omega_{mk}^{-1} )$.
Similarly, the $\tau$'s are approximated by Gamma factors in the conjugate exponential family,	for example
$q(\tau_u) = \Gcal(\tau_u ; \phi_u, \varphi_u)$.

The remaining question is, what to do with $p(\H)$, and the posterior marginal approximation $q(\H)$?

\section{Random graphs} \label{sec:randomgraphs}

Although an observation $g_{mn} = 1$ implies that $q(h_{mn} = 1) = 1$, we cannot estimate every one of $MN$ $q(h_{mn})$'s, as there are typically $10^{12}$ or more of them. 
As a recourse, we shall specify $q$ as an algorithm that stochastically generates connections $h_{mn} = 1$, so that $p(\H)$ produces (roughly) the same type of graphs as is observed in $\G$.

The graphical model in Figure \ref{fig:graphicalmodel} specifies that every ``considered'' edge $(m,n)$ in $\H$ contains a ``like'' probability $\sigma_{mn}$. For each edge in $\H$, a coin is flipped, and revealed with probability $\sigma_{mn}$ to give $\G$.
If we assume that the coin is on average unbiased, half the edges will be revealed, and $|\H| \approx 2|\G|$. Alternatively, $\G$ is a subnet of $\H$, containing half (or some rate of) its connections.
Working back, we sample graphs $\H$ at this rate, and the family of graphs $\H$ that can be generated this way constitutes our prior.
This places no guarantee that the two graphs will always be of the same type, as not all graph types are closed under random sampling.
For example, random subnets drawn from exact scale-free networks are not themselves scale-free \cite{Stumpf_2005}. However, the practical benefits of this algorithmic simplification  outweigh the cost of more exact procedures.

\subsection{Sampling $q(\H)$}

The factor $q(\H)$ is defined stochastically, with the criteria that it should not be too expensive to draw random samples $\H$. One approach would be to generate samples, similar to Section \ref{sec:realworld}, by specifying a degree distribution conditioned on the number of degrees $d$ that each user and item vertex in $\G$ has. If the mean of each is $2d$, one can show that a version of (\ref{eq:consistency}) will also hold for $\H$. At the cost of many redraws, one can sample half-edges, as in Section \ref{sec:realworld}, and connect them until all half-edges are paired.

We propose a simpler scheme here, which samples $\H$ from $\G$ in $\Ocal( |\G| \log N)$ time.
The scheme has the flavour of ``sampling by popularity'' \cite{Zeno_BPR_JMLR}.
We define a multinomial histogram ${\Mcal}(\bpi)$ on the $N$ items, where $\pi_n \ge 0$ for $n = 1, \ldots, N$. This mimics a pseudo degree distribution for missing degrees.
Let user $m$ have degree $d_m$, or have viewed $d_m$ items.
For user $m$, the subset of $d_m$ edges in $\H$ that corresponds to $g_{mn} = 1$ is marked.
We then sample $d_{m}$ random ``negative'' edges from ${\Mcal}(\bpi)$ \emph{without replacement}---this fills in the remaining values for row $m$ in $\H$, i.e.~$h_{mn}$ for $n = 1, \ldots, N$.
For user $m$ the sample \emph{without} replacement can be drawn in $\Ocal(\log N)$ time by doing bookkeeping with a weighed binary tree on the items.

There are many ways to define histogram $\bpi$, one of which is to simply let $\pi_n = d_n$, the number of degrees (or views) of item $n$. This is effectively a uniform prior: each item should have the same rate of negatives. If we believe that there is some quality bar that drives popular items to be more generally liked, the histogram can be adjusted with
\begin{equation} \label{eq:hist}
\pi_n = d_n^{\,\gamma}
\end{equation}
so that it obeys a version of the observed power law.
A free rate parameter $r$ is introduced, so that the most popular item with degree $d_{\max} = \max \{d_n\}$
has histogram weight
\begin{equation}
\pi_{\max} = r d_{\max} \ .
\end{equation}
As an example, $r = \frac{1}{2}$ will add half as many unobserved edges to $\H$ for that item. A substitution gives a power 
\begin{equation} \label{eq:adjustwithr}
\gamma = 1 + \log r / \log d_{\max}
\end{equation}
with which the histogram is adjusted in (\ref{eq:hist}).

\begin{figure}[t]
\begin{center}
\ifnum\figures=1
\includegraphics[width=0.45\textwidth]{Results/Figures/PositivesToNegatives.eps}
\else
\includegraphics[width=0.45\textwidth]{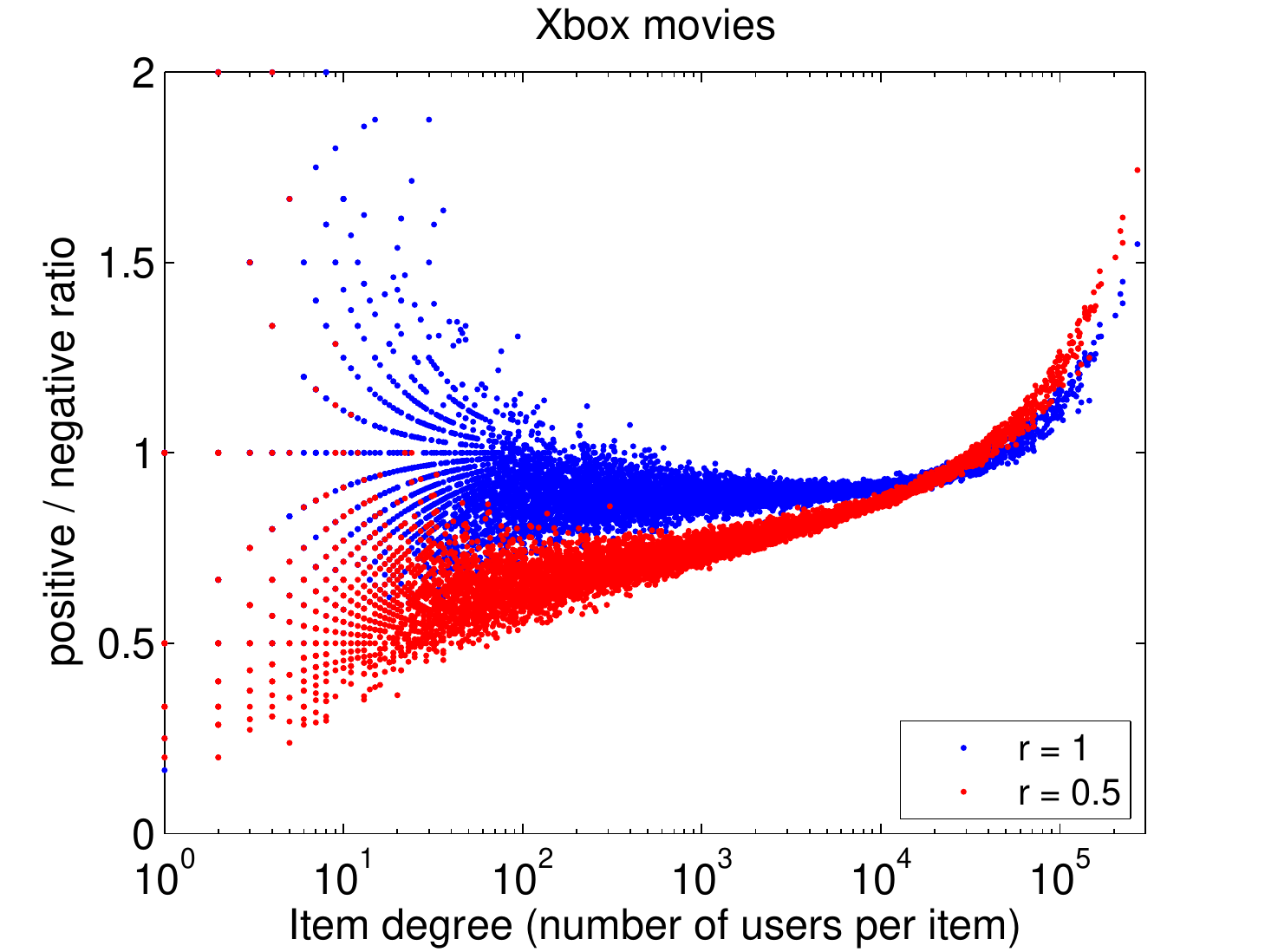}
\fi
\end{center}
\caption{The ratio of positive to negative edges per item, from a single sample from $q(\H)$.
(The ratio is skewed at the head: sampled edges to more popular items have higher odds to already exist in $\G$.
Discarding and resampling them leaves popular items underrepresented in the ``negative'' set. This can be overcome with another adjustment of $\bpi$ in $\Mcal(\bpi)$.)
}
\label{fig:pos-to-neg}
\end{figure}

Figure \ref{fig:pos-to-neg} shows two samples of the edges of $\H$ for two settings of $r$. For each item, it shows the ratio of ``positive'' to ``negative'' edges. A side effect is that at the head, the most popular items are underrepresented in the remainder of $\H$. This is because the items (or edges) sampled from ${\Mcal}(\bpi)$ might already exist in $\G$, and are discarded and another edge sampled.


\section{Variational inference} \label{sec:variational}

The approximation $q(\btheta)$ in (\ref{eq:factorizing}) is found by maximizing a variational lower bound on the  partition function of (\ref{eq:jointdensity}), with
\begin{equation} \label{eq:vbbound}
\log p(\G) \ge \Lcal[q] = \int q(\btheta) \log p(\G, \btheta) \, \mathrm{d} \btheta + \Hcal[q(\btheta)] \ .
\end{equation}
Here $\Hcal[q]$ is the (continuous) entropy of our choice of $q$.
The expression in (\ref{eq:vbbound}) is not analytically tractable due to the sigmoids in $\ell_{mn}$, which appear in $p(\G, \btheta)$ in (\ref{eq:jointdensity}), as they are not conjugate with respect to the $q(u_{mk})$'s or any of the other vertex factors.
We additionally lower-bound $\ell_{mn}$ with the logistic or Jaakkola-Jordan bound \cite{Jaakkola_Jordan_1996}, introducing an additional variational parameter $\xi_{mn}$ on each edge. The logistic bound is
\begin{equation} \label{eq:jjbound}
\ell \ge \mathrm{e}^{ g (\u^T \v + b) } 
\Big[ \sigma(\xi) \, \mathrm{e}^{ - \frac{1}{2} (\u^T \v + b + \xi )  
- \lambda(\xi) ( (\u^T \v + b)^2 - \xi^2 ) } \Big]^{g + h(1 - g)} ,
\end{equation}
where subscripts $m$ and $n$ that are clear from the context are suppressed. The bound depends on a deterministic function $\lambda(\xi) \defined \frac{1}{2 \xi} [ \sigma(\xi) - \frac{1}{2}]$.
The substitution of the lower bound in (\ref{eq:jjbound}) to $\ell_{mn}$ creates a  
$p_{\bxi}(\G, \btheta)$ that leaves the bounded likelihood conjugate with respect to its prior.
The bound $\Lcal_{\bxi}$,
\begin{equation} \label{eq:bound}
\Lcal[q] \ge \Lcal_{\bxi}[q] = \int q(\btheta) \log p_{\bxi}(\G, \btheta) \, \mathrm{d} \btheta + \Hcal[q]  \ ,
\end{equation}
is therefore explicitly maximized
over both the (variational) distribution $q$ and the additional variational parameters $\bxi = \{ \xi_{mn} \}$.

\bluetext{
\noindent\rule{\linewidth}{0.4pt}
\pointer
The bound in (\ref{eq:bound}) follows from
\begin{align*}
\log p(\G) & = \log \int p(\G, \btheta) \, \mathrm{d} \btheta \\
& \ge \log \int p_{\bxi}(\G, \btheta) \, \mathrm{d} \btheta \\
& = \log \int q(\btheta) \frac{p_{\bxi} (\G, \btheta)}{q(\btheta)} \, \mathrm{d} \btheta \\
& \ge \int q(\btheta) \log \frac{p_{\bxi} (\G, \btheta)}{q(\btheta)} \, \mathrm{d} \btheta
= \Lcal_{\bxi}[q] \ ,
\end{align*}
where $\Hcal[q] \defined - \int q(\btheta) \log q(\btheta) \, \mathrm{d} \btheta$.
The last line above follows from Jensen's inequality.
The quantity $\log p_{\bxi} (\G, \btheta)$ depends on a double sum over $MN$ plausible connections between users and items,
\begin{align*}
\log p_{\bxi} & 
= \sum_{m = 1}^{M} \sum_{n=1}^{N} \Bigg\{
g_{mn} \left(\u_m^T \v_n + b_{mn} \right) \\
&
\quad + \big( g_{mn} + h_{mn}(1 - g_{mn})\big) \Bigg[
\log \sigma(\xi_{mn}) \\
& \quad - \frac{1}{2} \left( \u_m^T \v_n + b_{mn} + \xi_{mn} \right) \\
& \quad - \lambda(\xi_{mn}) \left( \left(\u_m^T \v_n + b_{mn}\right)^2 - \xi_{mn}^2 \right) \Bigg] \Bigg\} \\
& \quad + \sum_{m=1}^{M} \left[ \log \Ncal(\u_m ; \0, \tau_u^{-1} \I) + \log \Ncal(b_m ; 0, \tau_{b_m}^{-1}) \right]
\\
& \quad + \sum_{n=1}^{N} \left[ \log \Ncal(\v_m ; \0, \tau_v^{-1} \I) + \log \Ncal(b_n ; 0, \tau_{b_n}^{-1}) \right] \\
& \quad + \log \Gcal(\tau_u ; \alpha, \beta) + \log \Gcal(\tau_v ; \alpha, \beta) \\
& \quad + \log \Gcal(\tau_{b_m} ; \alpha, \beta) + \log \Gcal(\tau_{b_n} ; \alpha, \beta) + \log  p(\H) \ ,
\end{align*}
and hence $\Lcal_{\bxi}[q]$ will also depend on $MN$ terms.
Again, the shorthand $b_{mn}$ is used for $b_n + b_m$.
Of course all the terms containing $h_{mn} = 0$ will drop away when 
$\log p_{\bxi}$ is evaluated. This is not the case for $\Lcal_{\bxi}$, as an expectation over those $h_{mn}$'s is required!

As a next step, the problem will be formulated
as a stochastic objective function that is ridden of this dependency.
Note that the dependence of $\Lcal_{\bxi}[q]$ on $\H$ is \emph{inside an expectation},
and $\log p_{\bxi}$ has a linear dependence on the $h_{mn}$ variables.
Since $q(\btheta)$ was chosen to factorize such that $q(\H)$ is a separate factor,
$\Lcal_{\bxi}$ can be rearranged in the form of a stochastic function over $q(\H)$.

For present purposes, split $\btheta$ up into $\{ \btheta_{\wo \H}, \H \}$, where ``$\wo$'' reads ``without''. Now $q(\btheta) = q(\btheta_{\wo \H}) \, q(\H)$.
As we've already stipulated a fixed scheme for generating samples from $q(\H)$,
the objective function can be written in terms of $q(\btheta_{\wo \H})$:
\begin{align*}
\Lcal_{\bxi} \big[ q(\btheta_{\wo \H}) \big] & 
= \mathbb{E}_{q(\btheta_{\wo \H})} \Bigg[ \sum_{m, n \, : \, h_{mn} = 1} \Bigg\{
\log \sigma(\xi_{mn})
\\
& \quad + \left( g_{mn} - \frac{1}{2} \right) \left(\u_m^T \v_n + b_{mn} \right) - \frac{1}{2} \xi_{mn}  \\
& \quad - \lambda(\xi_{mn}) \left( \left(\u_m^T \v_n + b_{mn}\right)^2 - \xi_{mn}^2 \right) \Bigg\} \\
& \quad + \sum_{m=1}^{M} \left[ \log \Ncal(\u_m ; \0, \tau_u^{-1} \I) + \log \Ncal(b_m ; 0, \tau_{b_m}^{-1}) \right]
\\
& \quad + \sum_{n=1}^{N} \left[ \log \Ncal(\v_m ; \0, \tau_v^{-1} \I) + \log \Ncal(b_n ; 0, \tau_{b_n}^{-1}) \right] \\
& \quad + \log \Gcal(\tau_u ; \alpha, \beta) + \log \Gcal(\tau_v ; \alpha, \beta) \\
& \quad + \log \Gcal(\tau_{b_m} ; \alpha, \beta) + \log \Gcal(\tau_{b_n} ; \alpha, \beta) 
\Bigg]  \\
& \quad + \Hcal \big[ q(\btheta_{\wo \H}) \big] \ .
\end{align*}
If we know $\H$, then the above formulation cuts the complexity from a sum over $MN$
terms to a sum over $|\H|$ terms, which is proportional to the observed graph's size.

Finally, we write $\Lcal_{\bxi}$ as a function that is stochastically dependent on $q(\H)$:
\[
\Lcal_{\bxi} [ q ] = 
\mathbb{E}_{q(\H)} \Bigg[ \Lcal_{\bxi} \big[ q(\btheta_{\wo \H}) \big] \Bigg] 
+ \Hcal \big[ q(\H) \big]\ .
\]
\emph{This formulation is key, and will allow us to do stochastic gradient updates of $\Lcal_{\bxi}$ over $q(\btheta_{\wo \H})$ and $\bxi$ by using random graphs generated from $q(\H)$.}

\vspace{-0.2cm} \noindent\rule{\linewidth}{0.4pt} \vspace{-0.2cm}
}

\subsection{Variational updates}

The variational updates for the user factors $q(u_{mk})$ are presented in this section. As the model is bilinear, the gradients of $\Lcal_{\bxi}$ with respect to the item factors can be set to zero following a similar pattern.
To minimize $\Lcal_{\bxi}$ with respect to $q(u_{mk})$, one might take functional derivatives
$\partial \Lcal_{\bxi} / \partial q(u_{mk})$ with respect to each $q(u_{mk})$, and sequentially equate them to zero.
This is slow, as each update will require a loop over all the vertex's edges: for the user, $K$ loops over all the items will be required.
The vertex factor can alternatively be updated in bulk, by first equating the gradients of $\Lcal_{\bxi}$ with respect to a full Gaussian (not factorized) approximation $\tilde{q}(\u_m)$ to zero.
The fully factorized $q(u_{mk})$  can then be recovered from the \emph{intermediate} approximation $\tilde{q}(\u_m)$ as those that minimize the Kullback-Leibler divergence $D_{\mathrm{KL}}( \prod_{k=1}^{K} q(u_{mk}) \| \tilde{q}(\u_m))$:
this is achieved when the means of $q(u_{mk})$ match that of $\tilde{q}(\u_m)$, while their \emph{precisions} match the diagonal precision of $\tilde{q}(\u_m)$.

How do we find $\tilde{q}(\u_m)$? The functional derivative $\partial \Lcal_{\bxi} / \partial \tilde{q}(\u_m)$ is zero where $\tilde{q}(\u_m)$ has as natural parameters a precision matrix of
\begin{equation} \label{eq:precisionupdate}
\P_m = \sum_{n=1}^{N} \Ebb_{q} \big[ h_{mn} \big] \cdot 2 \lambda(\xi_{mn}) \cdot \Ebb_{q} \big[ \v_n \v_n^T \big] + \Ebb_{q}[\tau_u] \I
\end{equation}
and mean-times-precision vector $\P_m \bmu_m$, which will be stated in (\ref{eq:update}).  Apart from having to average $h_{mn}$ over $q(\H)$, which we cannot do analytically, the update in (\ref{eq:precisionupdate}) suffers from having a summation over all $N$ item vertices.

The burden of having to determine a sum over a full item catalogue in (\ref{eq:precisionupdate}) can be removed with a  clever rearrangement of expectations.
As $h_{mn}$ is binary,
\begin{align}
\sum_{n=1}^{N} \Ebb_q \big[ h_{mn} \big] \, f(\v_n)
& = \sum_{\H} q(\H) \sum_{n=1}^{N} h_{mn} \, f(\v_n) \nonumber \\
& = \sum_{\H} q(\H) \sum_{n : h_{mn} = 1} f(\v_n) \ .
\label{eq:identity}
\end{align}
The sum over $\H$ in (\ref{eq:identity}) runs over all $2^{MN}$ possible instantiations of $\H$.
A rearrangement of (\ref{eq:precisionupdate}) therefore allows the updates to appear as a \emph{stochastic average},
\begin{align}
\P_m & = \Ebb_{q(\H)} \left[ \sum_{n : h_{mn} = 1} 2 \lambda(\xi_{mn}) \cdot \Ebb_{q} \big[ \v_n \v_n^T \big] + \Ebb_{q}[\tau_u] \I \right] \nonumber
\\
\P_m \bmu_m & = \Ebb_{q(\H)} \Bigg[ \sum_{n : h_{mn} = 1} \big(g_{mn} - \frac{1}{2} \cdots  \nonumber
\\
&  \qquad  \cdots  - 2 \lambda(\xi_{mn}) \cdot \Ebb_{q} \big[ b_m + b_n \big] \big) \Ebb_{q} \big[ \v_n \big] \Bigg] \ . \label{eq:update}
\end{align}
Inside the expectation over $q(\H)$, the mean field update in (\ref{eq:update}) 
is a quantity specified on the hidden graph $\H$ only, and not all $N$ plausible edges for the user.
We are able to sample graphs from $q(\H)$ according to Section \ref{sec:randomgraphs}.
Retrospectively, this choice now bears fruit,
as the update exists as an average amenable to stochastic gradient descent. 
We remark, too, that the natural parameters in (\ref{eq:update}) define the \emph{natural gradients} of the variational objective function \cite{Amari_1998,Sato_2001}.

\bluetext{
\noindent\rule{\linewidth}{0.4pt}
\pointer
We refer readers who require further insight into the role of natural parameters and gradients in stochastic variational inference to Hoffman et al.~\cite{JMLR:hoffman13a}, which was published in the same month as this paper. 

\vspace{-0.2cm} \noindent\rule{\linewidth}{0.4pt} \vspace{-0.2cm}
}

The full natural gradient is periodic in the number of vertices and the updates are component-wise, and convergence with such updates can also be achieved using a stochastic gradient algorithm \cite{Kushner_2003}.

There are additional variational parameters at play in  (\ref{eq:update}). For the required edges $h_{mn} = 1$ that connect user $m$ with items $n$, the values $\xi_{mn}$ that maximize $\Lcal_{\bxi}$ or $\Ebb_{q}[ \log p_{\bxi}(\G, \btheta)]$ are each given by
\begin{equation} \label{eq:logisticbound}
\xi_{mn}^{2} = \Ebb_{q} \big[ (\u_m^T \v_n + b_m + b_n)^2 \big] \ ,
\end{equation}
and they are computed and discarded when needed. We take the positive root as $\xi_{mn}$, and refer the reader to Bishop \cite{Bishop_2006} for a deeper discussion.

Given $\P_m$ and $\P_m \bmu_m$ from (\ref{eq:update}),
we have sufficient statistics for $\tilde{q}(\u_m)$, and hence for updating each of the $K$ $q(u_{mk})$'s in bulk. 
Deriving sufficient statistics for $q(v_{nk})$, $q(b_{m})$ and $q(b_n)$ is similar to that presented in (\ref{eq:update}), and the derivation will not be repeated.
Given these, optimization proceeds as follows: At time $t$, we sample a hidden graph $\H$, over which the user and item vertex factors are updated.
Focussing on user $m$, let $\P_m^{(t-1)}$ be the (diagonal) precision matrix of the factorized distribution $\prod_{k=1}^{K} q(u_{mk})$. We then find $\P_m$ in (\ref{eq:update}), and now the precision matrix of $\tilde{q}(\u_m)$ will be $\P_m^{(t)}$, found through $\P_m^{(t)} = \epsilon_t \P_m + (1 - \epsilon_t) \P_m^{(t-1)}$, where $\epsilon_t \in [0,1]$.
The mean-times-precision vector of $\tilde{q}(\u_m)$ is given through a similar stochastic update.

\bluetext{
\noindent\rule{\linewidth}{0.4pt}
\pointer
In particular, the loop over updates for users $m = 1, \ldots, M$ (which can be done in parallel) and items $n = 1, \ldots, N$ (which is also embarrassingly parallel) at time $t$ is \emph{preceded} by drawing a random sample $\H^{(t)}$ from $q(\H)$.

We'll focus on one update from the user-loop, as updates in the item-loop follow a mirrored form.
By $\P_m \bmu_m$, we imply the mean-times-precision vector, from which the mean can be solved. To make the distinction clear that it is a vector, we'll use $\z_m \defined \P_m \bmu_m$ below, such that $\bmu_m = \P_m^{-1} \z_m$.
The precise form of the update of the $q(u_{mk})$'s for user $m$ is then
\begin{align*}
\P_m & = \sum_{n : h_{mn}^{(t)} = 1} 2 \lambda(\xi_{mn}) \cdot \Ebb_{q} \big[ \v_n \v_n^T \big] + \Ebb_{q}[\tau_u] \I \\
\z_m & = \sum_{n : h_{mn}^{(t)} = 1} \big(g_{mn} - \frac{1}{2} \cdots
\\
&  \qquad  \cdots  - 2 \lambda(\xi_{mn}) \cdot \Ebb_{q} \big[ b_m + b_n \big] \big) \Ebb_{q} \big[ \v_n \big]  \ . 
\end{align*}
We now set
\begin{align}
\P_m^{(t)} & = \epsilon_t \P_m + (1 - \epsilon_t) \P_m^{(t-1)} \nonumber \\
\z_m^{(t)} & = \epsilon_t \z_m + (1 - \epsilon_t) \z_m^{(t-1)} \label{eq:alguserupdate} 
\end{align}
to give the natural parameters of $\tilde{q}(\u_m)$ at time $t$.

\vspace{-0.2cm} \noindent\rule{\linewidth}{0.4pt} \vspace{-0.2cm}
}

The factors $q(u_{mk})$ are then recovered from the bulk computation of $\tilde{q}(\u_m)$.

\bluetext{
\noindent\rule{\linewidth}{0.4pt}
\pointer
As $q(u_{mk}) = \Ncal(u_{mk} ; \eta_{mk}, \omega_{mk}^{-1})$, this recovery is achieved 
by first solving a linear system for the mean parameters $\bmu_m^{(t)} =  [ \P_m^{(t)} ]^{-1} \z_m^{(t)}$ (done stably by back solving the linear system twice using the Cholesky decomposition of $\P_m^{(t)}$). 
After $\bmu_m^{(t)}$ is obtained, set
\begin{equation} \label{eq:intermediate}
\eta_{mk} = \mu_{mk}^{(t)} \quad \textrm{and} \quad \omega_{mk} = [ \P_m^{(t)} ]_{kk} \ ,
\end{equation}
where the last subscript indicates the diagonal element $(k,k)$ of the precision matrix.
Note that we could have chosen the user factor to be a full Gaussian, with a $K \times K$ covariance matrix. In that case this step is not needed.

Solving for $\bmu_m^{(t)}$ is $\Ocal(K^3)$, which dominates an update of $\prod_{k=1}^{K} q(u_{mk})$ if $|\{ n : h_{mn}^{(t)} = 1\}|$ is small. In that case one may take partial gradients with respect to $\kappa < K$ components, with an $\Ocal(\kappa^3)$ inversion, and only update the subset of factors. This can be repeated until all $q(u_{mk})$ are updated.

The stochastic gradient step of the user biases $q(b_m)$ is similarly obtained in terms of its natural parameters, giving
\begin{align*}
P_m & = \sum_{n : h_{mn}^{(t)} = 1} 2 \lambda(\xi_{mn}) + \Ebb_{q}[\tau_{b_u}] \\
z_m & = \sum_{n : h_{mn}^{(t)} = 1} \big(g_{mn} - \frac{1}{2} 
- 2 \lambda(\xi_{mn}) \cdot \Ebb_{q} \big[ \u_m^T \v_n + b_n \big] \big)  \ . 
\end{align*}
(For simplicity, the notation for the natural parameters are overloaded above.) The updated precision and mean-times-precision parameters of $q(b_m)$ at time $t$ are therefore given by
\begin{align}
P_m^{(t)} & = \epsilon_t P_m + (1 - \epsilon_t) P_m^{(t-1)} \nonumber \\
z_m^{(t)} & = \epsilon_t z_m + (1 - \epsilon_t) z_m^{(t-1)} \ . \label{eq:algupdate} 
\end{align}

\vspace{-0.2cm} \noindent\rule{\linewidth}{0.4pt} \vspace{-0.2cm}
}

The series $\{ \epsilon_t \}_{t=1}^{\infty}$ should satisfy  $\sum_{t=1}^{\infty} \epsilon_t = \infty$ and  $\sum_{t=1}^{\infty} \epsilon_t^2 < \infty$, guarding against premature convergence and infinite oscillation around the maximum \cite{Robbins_Monro_1951}.

\bluetext{
\noindent\rule{\linewidth}{0.4pt}
\pointer
To avoid early local maxima (the problem is not convex), better results can be achieved by keeping $\epsilon_t = 1$ for the initial (say first $t_{\epsilon} = 10$) iterations.

\vspace{-0.2cm} \noindent\rule{\linewidth}{0.4pt} \vspace{-0.2cm}
}

Finally, the marginal approximations for the hyperparameters are updated by setting the functional derivatives, say $\partial \Lcal_{\bxi} / \partial q(\tau_u)$, to zero. For instance for
$q(\tau_u) = \Gcal(\tau_u ; \phi_u, \varphi_u)$ the shape $\phi_u$ and rate $\varphi_u$ are
\begin{align}
\phi_u & = \alpha + K M / 2  \nonumber \\
\varphi_u & = \beta + \frac{1}{2} \sum\nolimits_{m=1}^{M} \Ebb_q \big[\u_m^T \u_m \big] \ . \label{eq:tauupdate}
\end{align}
As $q(u_{mk})$ is dependent on $\H$, the rate is also stochastically updated as described above.

\bluetext{
\noindent\rule{\linewidth}{0.4pt}
\pointer
An algorithmic outline is provided in Algorithm \ref{alg}. Loops indicated with \textbf{\emph{p}for} are parallel for-loops.

\vspace{-0.2cm} \noindent\rule{\linewidth}{0.4pt} \vspace{-0.2cm}
}

\renewcommand*{\algorithmcfname}{\pointer Algorithm}
\begin{algorithm}[t]
\DontPrintSemicolon
\SetNlSty{}{}{:}
\SetAlgoNlRelativeSize{-3}
\SetKwFor{While}{\emph{p}for}{do}{end\emph{p}for} 
\bluetext{
\textbf{input:} $\G$, $K$, $\alpha$, $\beta$ \;
accumulator $a \leftarrow 0$ \;
step size $\epsilon \leftarrow 1$ \;
\For{$t=1 : t_{\max}$}{
sample $\H^{(t)} \sim q(\H)$ \;
\While{$m=1 : M$}{
update $q(b_m)$ using (\ref{eq:algupdate})\;
}
\While{$n=1 : N$}{
update $q(b_n)$, similar to (\ref{eq:algupdate}) \;
}
\While{$m=1 : M$}{
update $\prod_{k=1}^{K} q(u_{mk})$ using (\ref{eq:alguserupdate}) and (\ref{eq:intermediate}) \;
}
\While{$n=1 : N$}{
update $\prod_{k=1}^{K} q(v_{nk})$, similar to (\ref{eq:alguserupdate}) and (\ref{eq:intermediate}) \;
}
\If{$t > t_{\tau}$}{
\emph{/$\!$/ avoiding early local solutions} \;
update $q(\tau_{b_{u}})$, $q(\tau_u)$, $q(\tau_{b_{v}})$, and $q(\tau_v)$, similar to (\ref{eq:tauupdate}) \;
}
\If{$\Delta_t \defined t - t_{\epsilon} > 0$}{
\emph{/$\!$/ $1 - \Delta_t^{-0.6} \to 1$ from below as $\Delta_t \to \infty$} \;
$a \leftarrow (1 - \Delta_t^{-0.6}) a + 1$ \;
$\epsilon \leftarrow 1 / a$ \;
}
}
\caption{Stochastic VB over random graphs\label{alg}}
}
\end{algorithm}

\subsection{Large scale inference}

The use of a bipartite graph ensures that variational updates are parallelizable. For instance, by keeping all $q(v_{nk})$, $q(b_n)$ and $q(b_m)$ fixed for the item and user vertices, the gradients  $\partial \Lcal_{\bxi} / \partial \tilde{q}(\u_m)$, and hence the stochastic updates resulting from  (\ref{eq:update}), have no mutual dependence.
Consequently, the loop over user vertex updates $m = 1 \ldots M$ is embarrassingly parallel; the same is true for other updates. This will not hold for more general graphs like those of social networks, though, where more involved logic will be required.

Due to the fact that a variational lower bound is optimized for, optimization can also be distributed across multiple machines, \emph{as long as the bound holds}. For example,
one might distribute the graph according to item vertices in blocks $\Bcal_{b}$, and iteratively optimize one block at a time, or optimize blocks concurrently (with embarrassingly parallel optimization \emph{inside} the blocks, as discussed earlier). In this example the sparse user-item graph (matrix) $\G$ is distributed such that \emph{all} observations for a set $\Bcal_a$ of items are co-located on the same machine.
The natural gradients for the users then distribute across machines, and can be written so that the dependence on the data blocks on various machines separates.
When optimizing using the item-wise data block $\Bcal_{a}$ on one machine, we write $\P_m$ in  (\ref{eq:update}) as
\begin{align}
& \P_m  = \Ebb_{q(\H)} \left[
\mathop{\sum_{n : h_{mn} = 1}}_{n \in \Bcal_a} 2 \lambda(\xi_{mn}) \cdot \Ebb_{q} \big[ \v_n \v_n^T \big]  \right. \cdots \nonumber \\
& \quad \left. + \sum_{b \neq a}
\overbrace{
\mathop{\sum_{n : h_{mn} = 1}}_{n \in \Bcal_b} 2 \lambda(\xi_{mn}) \cdot \Ebb_{q} \big[ \v_n \v_n^T \big]
}^{\textrm{block $b$'s natural gradient } \X_{m}^{(b)} \textrm{; fixed}}
\right] + \Ebb_q[ \tau_u ] \I \ . \label{eq:blocks}
\end{align}
Update (\ref{eq:blocks}) defines a thin \emph{message interface} between various machines, where
each block has to communicate only its \emph{natural gradients} $\X_{m}^{(b)}$---and similar mean-times-precision gradients---to other blocks.\footnote{The division of data to machines will be dictated by the size of $M$ and $N$; for $N \ll M$ a user-wise division gives a smaller message interface, as only natural gradients for the items' updates will be required.}
In block $\Bcal_a$ we might iterate between updates (\ref{eq:blocks}) and full item updates for all $n \in \Bcal_a$, whilst keeping the incoming messages $\X_{m}^{(b)}$ from other machines fixed. After a few loops over users and items, one can move to the next block. Similarly, different machines can optimize on all the blocks $\{ \Bcal_b \}$ in parallel, as long as the natural gradient messages are periodically communicated to other machines.
The scheme presented here generalizes to a further subdivision of user vertices into blocks.


\section{Results} \label{sec:results}

Given $\G$, a pivotal task of collaborative filtering is that of accurately predicting the future presence of an edge. 
This allows online systems to personalize towards a user's taste by recommending items that the user might like.

The collaborative filtering model in Section \ref{sec:coll} explicitly separated the probability of a user considering an item from $\sigma$, the probability for the user liking the item.
The odds of liking an item depends on our inferred certainty of the user and item parameters,\footnote{We suppress
subscripts $m$ and $n$ for clarity, and write $q(\u)$ for the diagonal Gaussian $\prod_{k=1}^{K} q(u_{mk})$.}
\begin{align}
& p(g = 1 \, | \, h = 1) \approx \iiint \sigma(\u^T \v + b) \, q(\u) \, q(\v) \, q(b) \, \mathrm{d} \u \, \mathrm{d} \v \, \mathrm{d} b \nonumber \\
& \approx \int \sigma(a) \, \Ncal (a \, ; \, \mu_a, \sigma_a^2 ) \, \mathrm{d} a \approx \sigma \left( \mu_a \, \big/ \sqrt{1 + \pi \sigma_a^2 / 8} \right) \ . \label{eq:prediction}
\end{align}
The random variable $a$ was defined as $a \defined \u^T \v + b$, with its density approximated with its first two moments under $q$, i.e.~$\mu_a \defined \Ebb_q[\u^T \v + b]$ and $\sigma_a^2 \defined \Ebb_q[(\u^T \v + b - \mu_a)^2]$. The final approximation of a logistic Gaussian integral follows from MacKay \cite{MacKay_1992}.

$\phantom{.}$

\subsection{Evaluation}

We evaluated our model by removing a test set from the Xbox movies and Netflix (4 and 5 stars) data sets.
The degree distributions for these data sets are presented in Figure \ref{fig:degreedistributions}. 
The training data $\G_{\mathsf{train}}$ was created by randomly removing one edge (or item) for each user from $\G$; the removed edges formed the test set.

A core challenge of {\em real world} collaborative filtering algorithms is to find a balance between popular recommendations and personalized content in a structured form. 
Based on our experience, a criteria of a good recommender is the ability to suggest non-trivial items that the user will like, and surface
less popular items in the tail of the item catalogue.
In the evaluations we highlight this by grouping results according to item popularity in Figure \ref{fig:averagerank}, for example.

Two evaluations are discussed below. Firstly, given that an item is presented to a user with $h_{mn} = 1$, we are interested in the classifying $g_{mn} \to \{ 0, 1\}$. This is one of the key contributions that our model brings to the table. As far as we are aware, there are no other algorithms that isolate $p(\mathrm{like})$ in this way.
To be able to draw a comparison with a 
known state-of-the-art algorithm, we consider various forms of a rank-based metric in a second evaluation.

In the tests below, $K=20$ latent dimensions were used. The user biases were clamped at zero, as $q(\H)$ was defined to give balanced samples for each user.
The rate and shape parameters of the hyperprior were set to $\alpha = \beta = 0.01$, giving a hyperprior on the $\tau$'s with mean 1 and a flexible variance of 100. The means of the hyperparameter posterior estimates were
$\Ebb[\tau_{b_v}] = 0.4$, $\Ebb[\tau_v] = 3.5$, and $\Ebb[\tau_u] = 2.0$.
When rounded to one decimal place, these values were similar on both the Netflix (4 and 5 stars) dataset and the Xbox Movies dataset.

\subsubsection{The ``like'' probability}

\begin{figure} [t]
\begin{center}
\ifnum\figures=1
\includegraphics[width=0.45\textwidth]{Results/Figures/ClassificationError.eps}
\else
\includegraphics[width=0.45\textwidth]{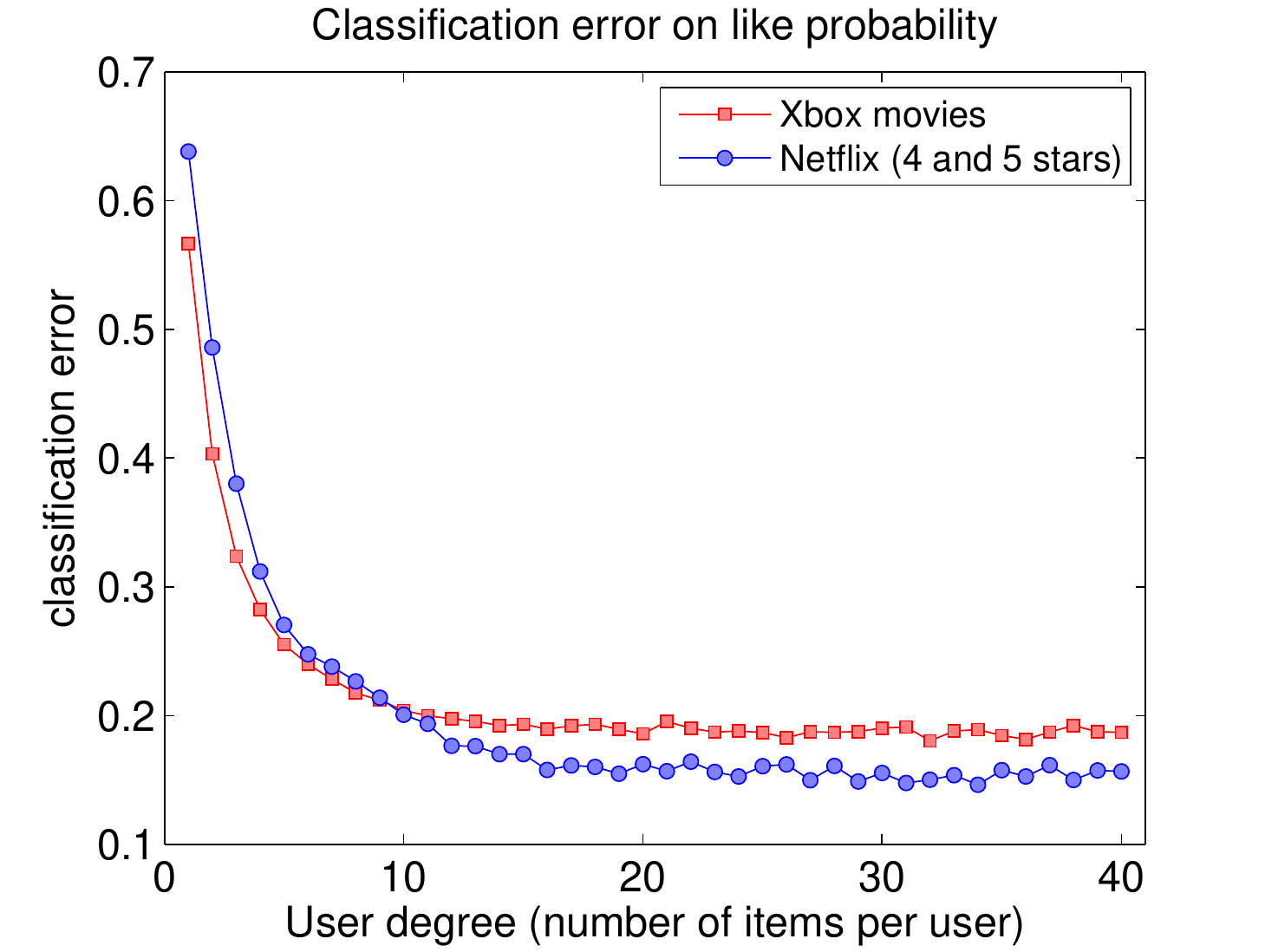}
\fi
\end{center}
\caption{The classification error on $\G_{\mathsf{test}}$, given $h = 1$ (the ground truth is $g = 1$).
The full histograms of probabilities $p(g = 1 | h = 1)$ are presented in Figure \ref{fig:probabilities}.
}
\label{fig:classificationerror}
\end{figure}

\begin{figure} [t]
\begin{center}
\ifnum\figures=1
\includegraphics[width=0.45\textwidth]{Results/Figures/XboxLike.eps} \\
\tiny{\whitetext{.}} \\
\includegraphics[width=0.45\textwidth]{Results/Figures/Netflix4and5Like.eps}
\else
\includegraphics[width=0.45\textwidth]{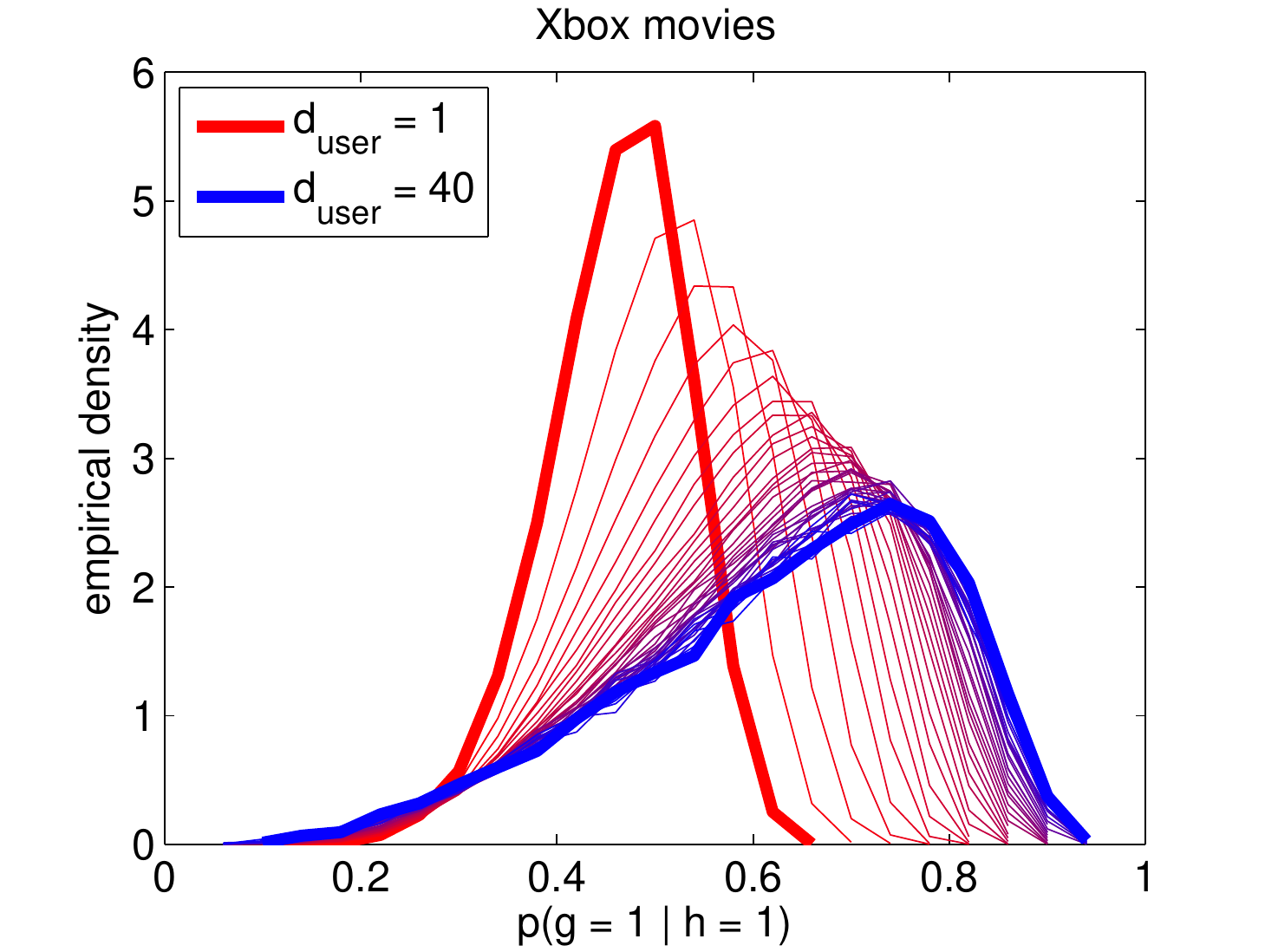} \\
\tiny{\whitetext{.}} \\
\includegraphics[width=0.45\textwidth]{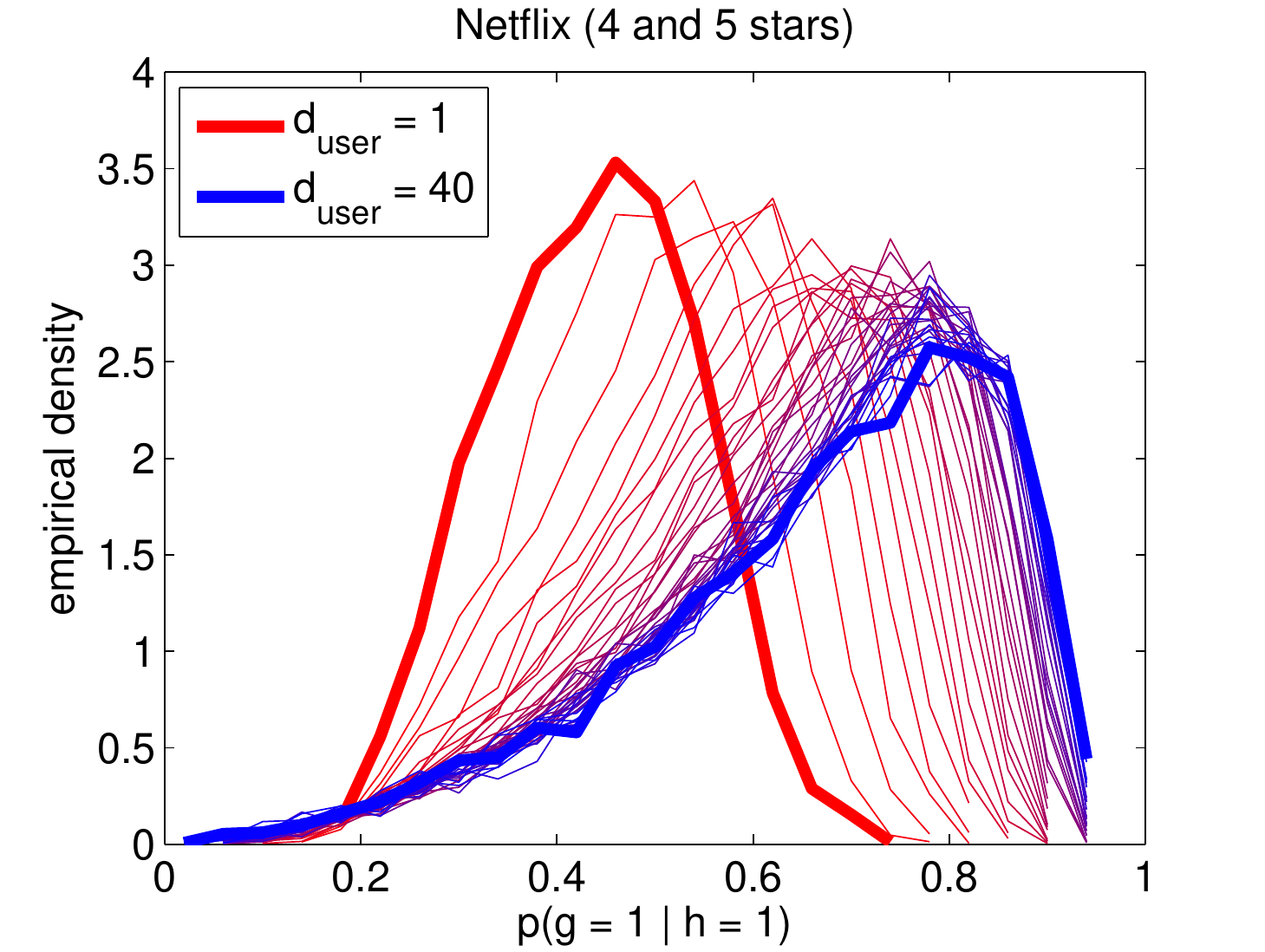}
\fi
\end{center}
\caption{The distribution of $p(g_{mn} = 1 | h_{mn} = 1)$ on the held out items in the evaluation, sliced incrementally according  to users
connected to $d_{\usr} = 1$ to 40 items. The ground truth is $g_{mn} = 1$.}
\label{fig:probabilities}
\end{figure}

The classification error on the held-out data converges to a stable value as users view between ten and twenty items.
Its plot is presented in Figure \ref{fig:classificationerror}, and has a natural interpretation.
Conventional wisdom dictates that the error rates for explicit ratings-based recommendation systems are typically in the order of 20\% of the ratings range. For Netflix's five-star ratings datasets, this error is around one star \cite{BellKorGrandPrize}, while an error of around 20 points in the 0-100 scale of the Yahoo! Music dataset is usual \cite{YahooMusic}.
The 16-19\% classification error in Figure \ref{fig:classificationerror} is therefore in line with the signal to noise ratio in well known explicit ratings datasets.
When users viewed only one item, the bulk of the predictive probability mass $p(g = 1 | h = 1)$ is centered around 50\%, slightly skewed to being less certain.
This is illustrated in Figure \ref{fig:probabilities}. As users view more items, the bulk of the predictive probability  skews towards being more certain\footnote{A property of a good probabilistic classification system is that it produces an exact \emph{callibration plot}. For example, we expect 10\% of edges to be misclassified for the slice of edges that are predicted with $p(g = 1 | h = 1) = 10\%$. The callibration plot requires a ground truth negative class $g = 0$, which is \emph{latent} in our case.
Figures \ref{fig:classificationerror} and \ref{fig:probabilities} aim to present an equivalent to a callibration plot.}.

The probability $p(g = 1 | h = 1)$ is useful in presenting a user with interesting recommendations, as it is agnostic to each item's popularity. It is therefore possible to define a \emph{utility function} that trades this quantity off with an item's popularity, effectively giving a knob to emphasize exploration or exploitation. Such a utility can be optimized through A/B tests in a flighting framework, but is beyond the scope of this paper.

\subsubsection{Average rank}

\begin{figure*} [t]
\begin{center}
\ifnum\figures=1
\includegraphics[width=0.45\textwidth]{Results/Figures/XboxUsers.eps}
\includegraphics[width=0.45\textwidth]{Results/Figures/XboxItems.eps} \\
\tiny{\whitetext{.}} \\
\includegraphics[width=0.45\textwidth]{Results/Figures/Netflix4and5Users.eps}
\includegraphics[width=0.45\textwidth]{Results/Figures/Netflix4and5Items.eps}
\else
\includegraphics[width=0.45\textwidth]{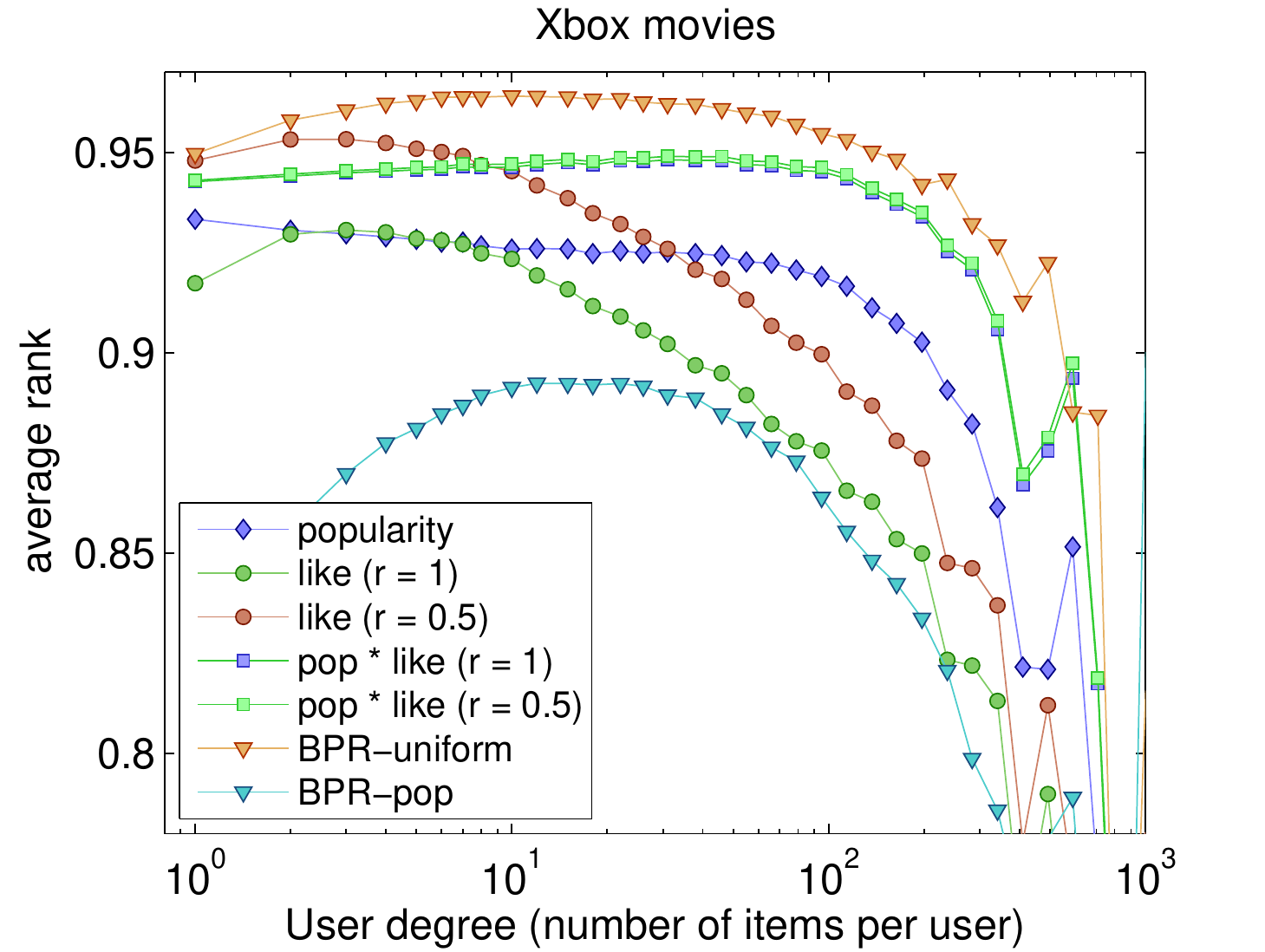}
\includegraphics[width=0.45\textwidth]{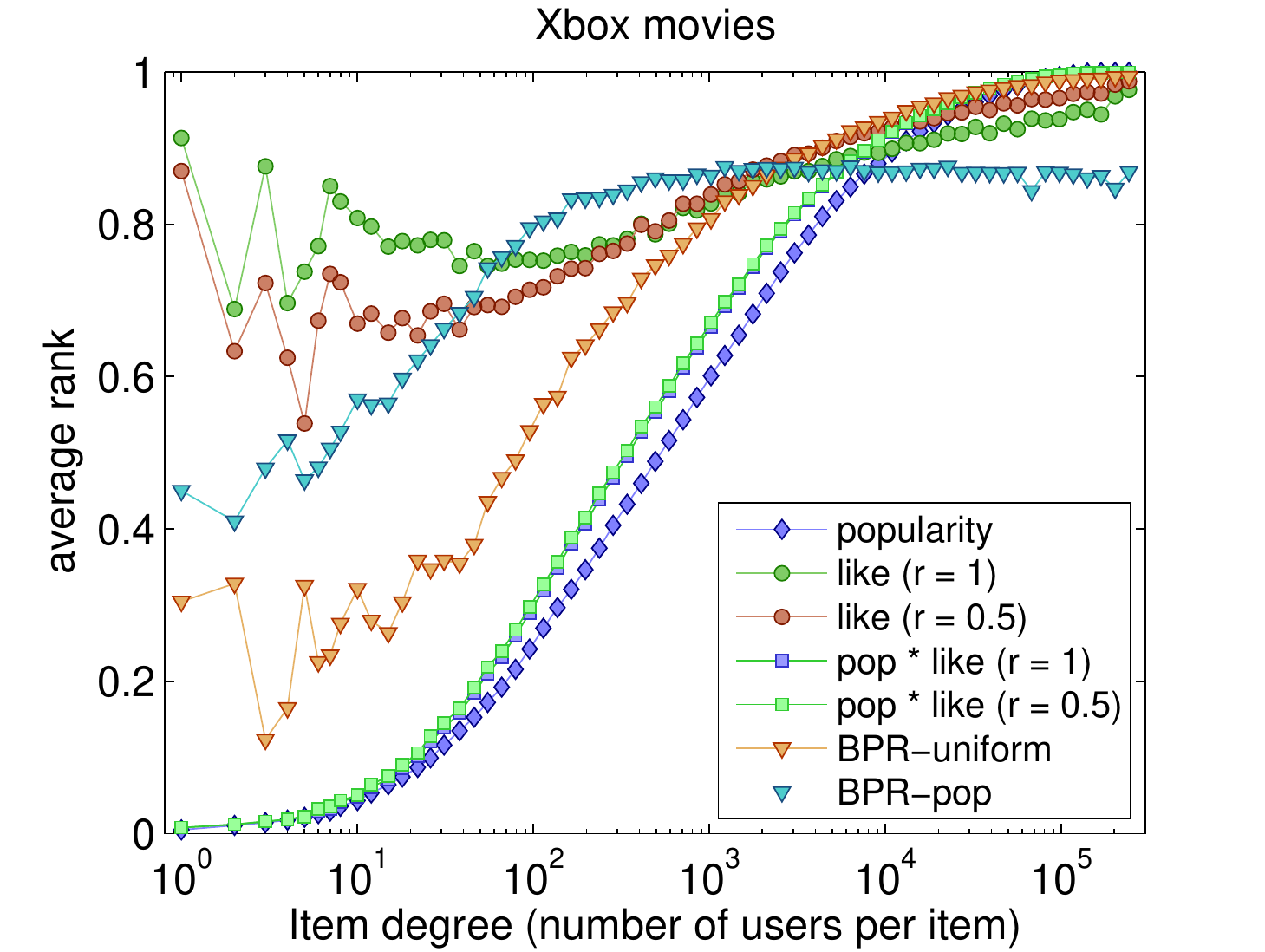} \\
\tiny{\whitetext{.}} \\
\includegraphics[width=0.45\textwidth]{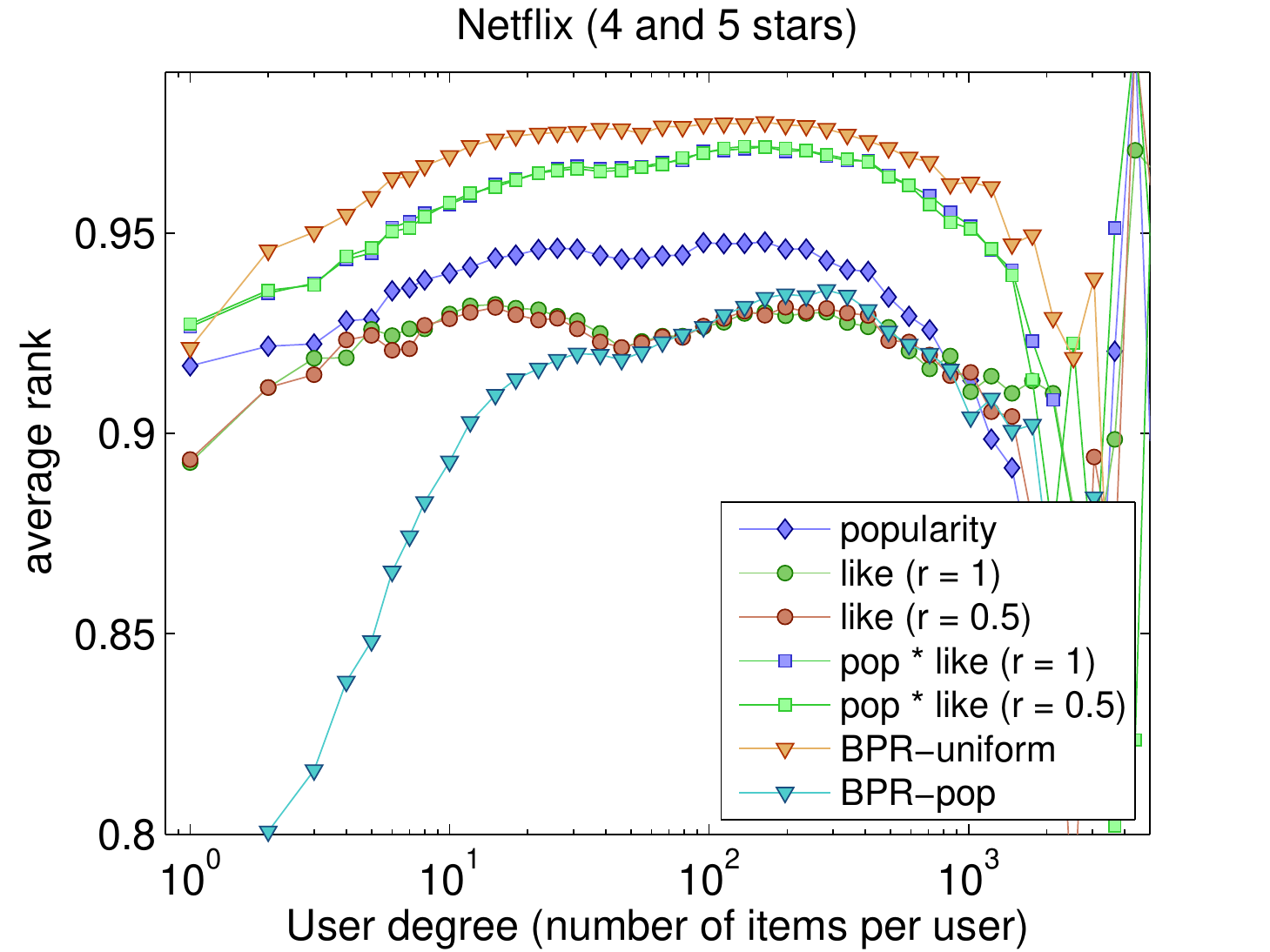}
\includegraphics[width=0.45\textwidth]{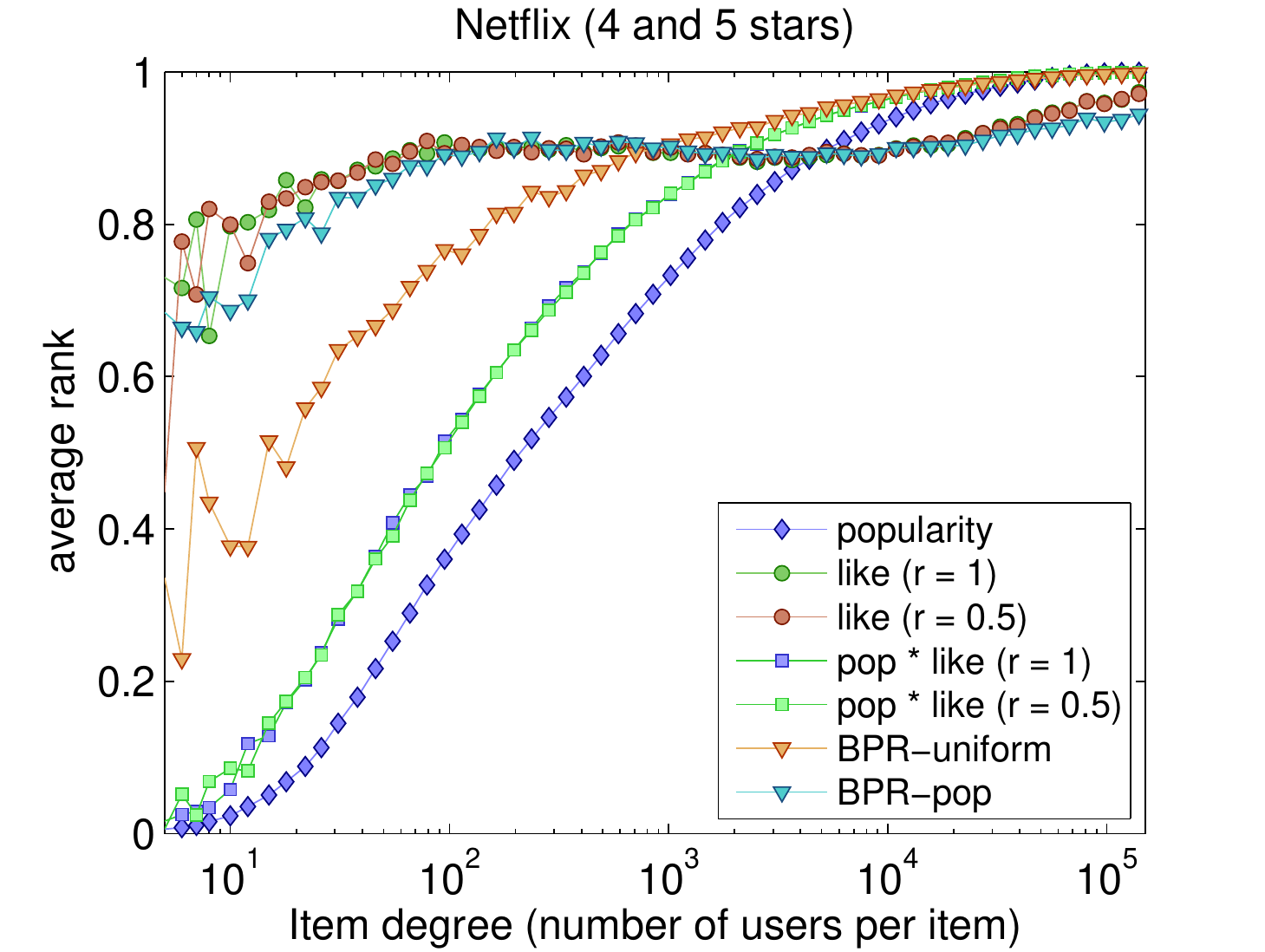}
\fi
\end{center}
\caption{The rank $S_{\mathrm{rank}}(m,n)$ in (\ref{eq:score}), averaged over users \emph{(left)} and items \emph{(right)}, grouped logarithmically by their degrees.
The \emph{top} evaluation is on the Xbox movies sample, while the \emph{bottom} evaluations are on the Netflix set, as given in Figure \ref{fig:degreedistributions}.
}
\label{fig:averagerank}
\end{figure*}

We turn to a ranking task to draw a comparison against known work, 
as we are unaware of other algorithms that isolate $p(\mathrm{like})$.
On seeing $\G_{\mathsf{train}}$, the absent edges (where $g_{mn} = 0$) are ranked for each user $m$. The ranking is based on various scores $s_{mn}$:
\begin{description}[font=\sffamily\bfseries] 
\item[like] the odds of a user liking an item, namely $s_{mn} = p(g_{mn} = 1 \, | \, h_{mn} = 1)$ as approximated in (\ref{eq:prediction});
\item[popularity] $s_{mn} = \pi_n$;
\item[popularity$\times$like] the odds of a user considering \emph{and} liking an item, namely $s_{mn} = \pi_n \,  p(g_{mn} = 1 \, | \, h_{mn} = 1)$.	
\end{description}
We evaluated models for the two settings of $r$ in (\ref{eq:adjustwithr});
a sample from $\H$ for each was shown in Figure \ref{fig:pos-to-neg}.

Our metric is computed as follows: If item $n'$ was removed, the rank score counts the position of $n'$ in an ordered prediction list
\begin{equation} \label{eq:score}
S_{\mathrm{rank}}(m,n') \defined \sum_{n : g_{mn} = 0} \Ibb \Big[ s_{mn'} > s_{mn} \Big]  \, \Bigg/  \sum_{n : g_{mn} = 0} 1 \ .
\end{equation}
Random guessing would give $S = 0.5$, while $S = 1$ places the held-out item at the head of the list.

As a benchmark, we use the Bayesian Personalized Ranking (BPR) model of Rendle et al.~\cite{MyMediaLite, Rendle_2009}. It has shown state of the art performance on ranking metrics against methods ranging from singular value decompositions and nearest neighbours to weighed regularized matrix factorization \cite{Pan_2008}.

BPR was also used as a key component in many of the leading solutions for the second track of the KDD-Cup'11 competition \cite{KDDCup11}.
The competition was designed to capture the ability of models to personalize recommendations that ``fit'' specific users regardless of an item's popularity.
In that setting, BPR was trained with missing items sampled with probabilities proportional to their popularity as described in \cite{Zeno_BPR_JMLR}.
We therefore implemented and trained two BPR models: 
\begin{description}[font=\sffamily\bfseries] 
\item[BPR-uniform] with missing items sampled uniformly;
\item[BPR-popularity] with missing items sampled proportional to their popularity.
\end{description}
These two models capture two different aspects of recommender systems. \textbf{\textsf{BPR-uniform}} is optimized to learn a user-wise ranking of items, where the objective function specifies that
items that are liked (i.e.~$g_{mn} = 1$) should be ranked above missing items (i.e.~$g_{mn} = 0$).

The metric in (\ref{eq:score}) follows \cite{Rendle_2009}. Because \textbf{\textsf{BPR-uniform}} directly optimizes this metric, it should come as no surprise that it will perform better than methods that do not optimize it directly (see Figure \ref{fig:averagerank}). However, meaningful insights can still be gleaned from the comparison.
\textbf{\textsf{BPR-popularity}} is aimed at ranking observed ``liked'' items  above other {\em popular} items that are missing from the user's history.
While two BPR models are required to capture these two different aspects of recommendations, our generative model captures both of these aspects in a structured manner.

Figure \ref{fig:averagerank} illustrates the mean rank scores, grouped logarithmically by user and item degrees. 
In the plots that are grouped by user degrees, we see improved results for algorithms that prefer popularity, i.e.~\textbf{\textsf{popularity$\times$like}} and \textbf{\textsf{BPR-uniform}}.
This is explained by the dominance of popularity biases in both datasets. 
As expected, \textbf{\textsf{BPR-uniform}} show best results as it is optimizes the task at hand directly.
The estimates for users with an order of $10^{3}$ to $10^{4}$ degrees are noisy
as the data is very sparse (see Figure  \ref{fig:degreedistributions}).
However, when looking at the per item breakdown, we learn that \textbf{\textsf{BPR-uniform}} and the
\textbf{\textsf{popularity$\times$like}}
models perform poorly on less popular items and their superior results are based on recommendations from the short head of the popular items. When it comes to recommending from the long tail of the less familiar items, the \textbf{\textsf{like}} models show best results, with \textbf{\textsf{BPR-popularity}} just behind. These trends are consistent on both datasets.

\begin{figure} [t]
\begin{center}
\ifnum\figures=1
\includegraphics[width=0.45\textwidth]{Results/Figures/XboxPercentilesLike.eps} \\
\tiny{\whitetext{.}} \\
\includegraphics[width=0.45\textwidth]{Results/Figures/XboxPercentilesPopLike.eps}
\else
\includegraphics[width=0.45\textwidth]{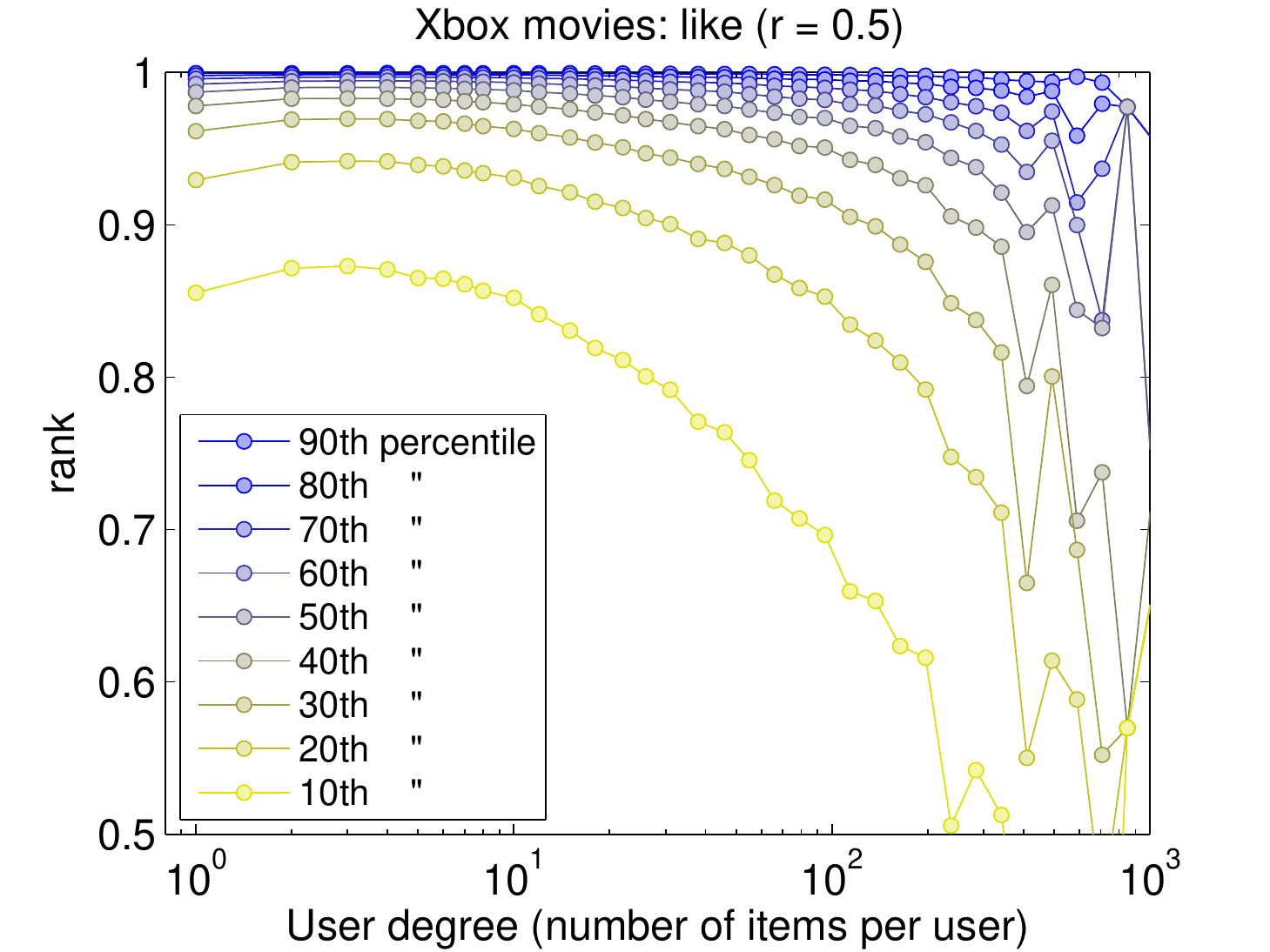} \\
\tiny{\whitetext{.}} \\
\includegraphics[width=0.45\textwidth]{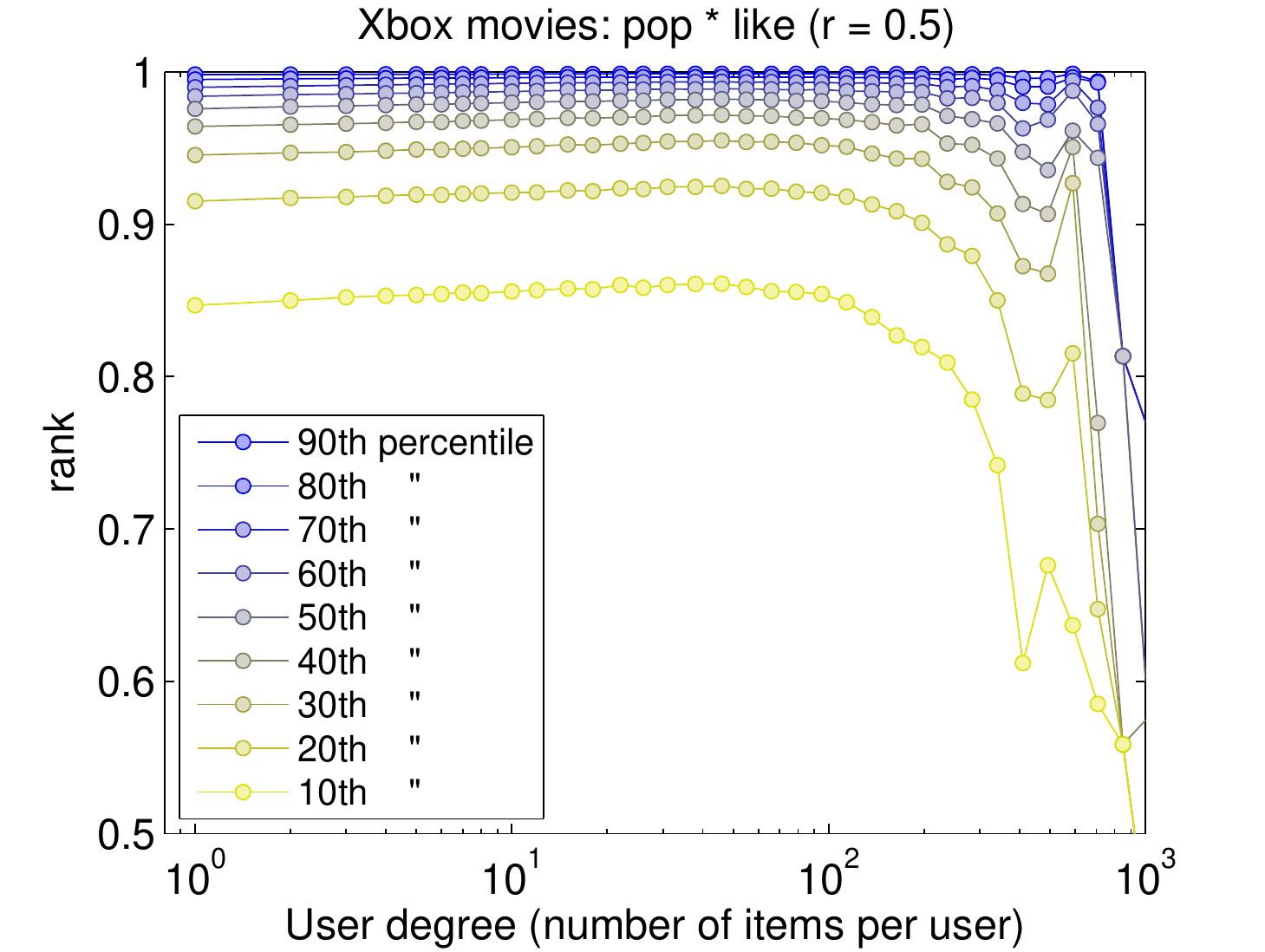}
\fi
\end{center}
\caption{Error bars on the rank tests. The median is much higher than the average rank reported in Figure \ref{fig:averagerank}.}
\label{fig:stats}
\end{figure}

The distribution of the ranks over all users (and items) is heavy-tailed, and whilst the average is often reported, the median is much higher than the average reported in Figure \ref{fig:averagerank}.
Figure \ref{fig:stats} shows the \emph{error bars} using the percentiles of the rank scores for tests \textbf{\textsf{like}} and \textbf{\textsf{popularity$\times$like}} for $r = \frac{1}{2}$.
The rank variance decreases as users view a few movies, but increases for heavy users which are harder to model. When popularity is included in the ranking, the error bars get tighter for heavy users, which implies that these users' lists are mostly governed by popularity patterns. 

\section{Conclusions}

Random graphs can be leveraged to predict the presence of edges in a collaborative filtering model.
In this paper we showed how to incorporate such graphs in an inference procedure by rewriting a variational Bayes algorithm in terms of random graph samples.
As a result, we were able to explicitly extract a ``like'' probability that is largely agnostic to the popularity of items.
The use of a bipartite graph, central to this exposition, is not a hindrance, as user-user interactions in a general network can be similarity modelled with $\sigma(\u_{m}^T \u_{m'})$.
While scalable parallel inference is not immediately obvious, we believe this to be a worthwhile pursuit.
By employing the same machinery on general graphs, one should be able to model connections in social or other similar networks. 

The use of a Bayesian graphical model makes it easy to adapt the model to incorporate richer feedback signals. Similarly, both structured and unstructured meta-data can be plugged into the graphical model.
The hidden graph $\H$ may also be partly observed, for example from system logs.
In that case some true negatives exist. Alternatively, we may know \emph{a priori} when a user could never have considered an item, fixing some $h$ at zero.
In both these scenarios the process of drawing random hidden graphs $\H$ can be adjusted accordingly.
For the sake of clarity, none of these enhancements were included in this paper.

\section{Acknowledgments}

The authors are indebted to Nir Nice, Shahar Keren, and Shimon Shlevich for their invaluable input, management, and stellar engineering skills.

\bibliographystyle{abbrv}
\bibliography{nips}

\end{document}